\documentclass[10pt,letterpaper]{article}
\usepackage[top=0.85in,left=2.35in,right=1in,footskip=0.75in]{geometry}

\usepackage{amsmath,amssymb}
\usepackage{bm}
\usepackage{dirtytalk}

\usepackage{siunitx}
\usepackage[T1]{fontenc}
\usepackage[utf8]{inputenc}
\usepackage[UKenglish]{babel}

\usepackage{changepage}

\usepackage{textcomp,marvosym}

\usepackage{cite}

\usepackage{nameref,hyperref}
\usepackage{cleveref}

\usepackage{microtype}
\DisableLigatures[f]{encoding = *, family = * }

\usepackage[table]{xcolor}
\usepackage{graphicx}
\usepackage{changes}
\usepackage{placeins}

\usepackage{array}

\newcolumntype{+}{!{\vrule width 2pt}}

\newlength\savedwidth

\usepackage{setspace} 
\usepackage[aboveskip=1pt,labelfont=bf,labelsep=period,justification=raggedright,singlelinecheck=off]{caption}

\makeatletter
\renewcommand{\@biblabel}[1]{\quad#1.}
\makeatother

\newcommand{\stgan}{\texttt{StyleGAN3}{}}

\graphicspath{{figures/},{data/}}

\usepackage{lastpage,fancyhdr,graphicx}
\usepackage{epstopdf}
\fancyheadoffset[L]{2.25in}
\fancyfootoffset[L]{2.25in}
\begin{document}
\vspace*{0.2in}

\begin{flushleft}
{\Large
\textbf{Synthetic images aid the recognition of human-made art forgeries} % Please use "sentence case" for title and headings (capitalize only the first word in a title (or heading), the first word in a subtitle (or subheading), and any proper nouns).
}
\newline
\\
Johann Ostmeyer\textsuperscript{1\P$\dagger$},
Ludovica Schaerf\textsuperscript{2\P},
Pavel Buividovich\textsuperscript{1*},
Tessa Charles\textsuperscript{3},
Eric Postma\textsuperscript{4}, 
Carina Popovici\textsuperscript{2}
\\
\bigskip
\textbf{1} Department of Mathematical Sciences,
	University of Liverpool, Liverpool L69 3BX, United Kingdom
\\
\textbf{2} Art Recognition AG,
	Soodmattenstrasse 4, CH-8134 Adliswil, Switzerland
\\
\textbf{3} Australian Synchrotron, Australian Nuclear Science and Technology Organisation (ANSTO), Clayton, 3168, Australia
\\
\textbf{4} Cognitive Science \& AI,
	Tilburg University, 5037 AB Tilburg, the Netherlands
\bigskip

% Insert additional author notes using the symbols described below. Insert symbol callouts after author names as necessary.
% 
% Remove or comment out the author notes below if they aren't used.
%
% Primary Equal Contribution Note
\P These authors contributed equally to this work.

% Use the asterisk to denote corresponding authorship and provide email address in note below.
$^\dagger$ \href{mailto:J.Ostmeyer@liverpool.ac.uk}{J.Ostmeyer@liverpool.ac.uk}\\
* \href{mailto:Pavel.Buividovich@liverpool.ac.uk}{Pavel.Buividovich@liverpool.ac.uk}

\end{flushleft}
\section*{Abstract}
Previous research has shown that Artificial Intelligence is capable of distinguishing between authentic paintings by a given artist and human-made forgeries with remarkable accuracy, provided sufficient training. However, with the limited amount of existing known forgeries, augmentation methods for forgery detection are highly desirable. 
In this work, we examine the potential of incorporating synthetic artworks into training datasets to enhance the performance of forgery detection. Our investigation focuses on paintings by Vincent van Gogh, for which we release the first dataset specialized for forgery detection. To reinforce our results, we conduct the same analyses on the artists Amedeo Modigliani and Raphael. We train a classifier to distinguish original artworks from forgeries. For this, we use human-made forgeries and imitations in the style of well-known artists and augment our training sets with images in a similar style generated by Stable Diffusion and StyleGAN. We find that the additional synthetic forgeries consistently improve the detection of human-made forgeries. In addition, we find that, in line with previous research, the inclusion of synthetic forgeries in the training also enables the detection of AI-generated forgeries, especially if created using a similar generator.

%\linenumbers
\section*{Introduction}

Forgeries are a serious threat to the artwork market, as illustrated for instance by the infamous Max Ernst forgery ``La Horde''. In 2006, the auction house Christie's announced the sale of the artwork, with an estimated value of about £3,000,000. However, it turned out that ``La Horde'' was a forgery created by the art forger Wolfgang Beltracchi~\cite{forgeriesGuardian}. Similarly, at the beginning of the 20th century, the Wacker case made the headlines globally. The German art dealer Otto Wacker, possibly with the help of his brother Leonhard, managed to sell over 30 fake Van Gogh paintings to public and private collectors, and many of the paintings were even included in the Catalogue Raisonné by Van Gogh expert Jacob de la Faille \cite{Bailey_2021}. Despite experts' disagreement, the art dealer was charged with fraud in April 1932. 

Recent developments in computer vision and machine learning techniques may contribute to the issue in several ways \cite{BellArtComputerVision}. 

Starting from late 1990s, various computer vision and image analysis techniques such as fractal analysis \cite{TaylorPollockFractalAnalysis99}, wavelets \cite{RockmoreAuthentication2004,Postma2008_art_classification,QiVisualStylometry2013}, sparse coding \cite{RockmoreSparseCoding2010}, clustering-based segmentation \cite{LiBrushstrokeExtraction2012} and tight frame method \cite{LiuTightFrame2016} were applied to extract characteristic features of individual artist's style automatically. More recently, the development of efficient classifier neural networks such as Convolutional Neural Networks (CNNs) allowed reaching very high accuracies in artwork attribution \cite{vanNoordPostmaCNNs,RasRijksmuseumCNNs,Zhu:1911.10091,CETINIC2018107,Pancaroglu:2012.01009,VARSHNEY2023105734}. 

While most of the studies concentrate on the attribution of an artwork to several pre-defined authors, similar machine-learning methods can also be used to distinguish between authentic artworks by a given author and forgeries. Due to the very close resemblance between original images and human-made forgeries (such as the Wacker forgeries), art authentication is generally a more challenging task than artwork attribution. In particular, authentication algorithms often have to learn very fine details such as brushstroke structure \cite{david_authentication_2021,QiVisualStylometry2013,LiBrushstrokeExtraction2012}. An additional challenge is that, for a given artist, forgeries are typically much less numerous than original artworks and often lack systematic documentation and high-resolution scans or photos. Despite such limitations, in recent years, NNs such as Convolutional Neural Networks (CNNs) or transformer-based architectures have shown promising results in both art attribution, when trained on datasets of authentic paintings and other stylistically similar artworks~\cite{david_authentication_2021} as well as in artwork authentication, trained against forgeries \cite{Schaerf2023}. 

In this context, the new trend of Generative Artificial Intelligence (GenAI) appears to present both threats and opportunities. On the one hand, GenAI might be adopted to create refined synthetic digital forgeries~\cite{Paint_it_Black}, which might populate the internet and diffuse misinformation. The possibility of creating `fake' synthetic artworks using AI-based methods gained popularity with the publication of Neural Style Transfer (NSF) \cite{gatys_image_2016}, which learns to decouple the style of an artwork from its content. This method is capable of creating synthetically styled images in the particular predisposition of a given artist to varying scales of accuracy and applicability. The successive publication of various Generative Adversarial Network enhanced architectures (e.g.\ StyleGANs~\cite{stylegan2,stylegan3}) and powerful large-scale diffusion models (e.g.\ Stable Diffusion~\cite{Rombach22} and DALL-E 2~\cite{Ramesh2021}) paved the way to the generation of realistic synthetic forgeries. In particular, the introduction of text conditioning using contrastive text-image models such as CLIP~\cite{CLIP_guidance}, created an accessible and quick interface for the creation of artworks `in the style of'. Differently from NSF, the latter is not tied to an input natural image and therefore allows greater freedom of generation.

At the same time, the ability of GenAI to create synthetic forgeries may mitigate the limitation of AI-based art authentication of being hampered by the limited availability of known forgeries and imitations. The goal of this work is to explore to what extent the recent GenAI methods such as StyleGANs and Stable Diffusion are able to augment the training datasets of known forgeries and enhance the performance of AI-based art authentication. While most recent proposals to detect fake images mainly address photorealistic images~\cite{Sha22}, the use of synthetic forgeries in artwork authentication is a widely unexplored area. Specifically, we focus on paintings by Vincent van Gogh, which are frequently used as a benchmark dataset for machine-based art attribution methods \cite{Postma2008_art_classification,QiVisualStylometry2013,LiBrushstrokeExtraction2012,LiuTightFrame2016}. Van Gogh painted a sheer amount of artworks, now in the public domain, and was widely forged due to its enormous market value. Van Gogh datasets, therefore, serve as valuable case studies for art authentication.

We build on the already publicly available dataset VGDB-2016 on Van Gogh~\cite{folego2016vangogh} available \href{https://figshare.com/articles/dataset/From\_Impressionism\_to\_Expressionism\_Automatically\_Identifying\_Van\_Gogh\_s\_Paintings/3370627}{here}, containing a set of $126$ RGB original images by the artist and a set of $212$ non-authentic RGB images by other Impressionist and Expressionist artists. The VGDB-2016 dataset does not contain forgeries, making it unsuitable for forgery detection. To address this, we enrich it for the purposes of art authentication and add $11$ RGB images of well-known forgeries created by the forger Otto Wacker into our dataset. We also include $8$ forgeries by former art forger and now legitimate artist creating \textit{genuine fakes}, John Myatt. The latter images are not in the Open Domain, therefore we only provide a pointer to those images. Furthermore, we release the artificially generated AI-based forgeries specifically generated for this paper. Finally, to reinforce our findings on van Gogh, we carry out the same analysis on datasets of Amedeo Modigliani and Raffaello Sanzio (Raphael), which are detailed in the supporting information \nameref{S1_Appendix}.

The outline of this paper is as follows. The next section details the \nameref{sec:methods} we employ to generate synthetic images used to augment the training data set of known forgeries. We also briefly discuss the classifier model that we use for forgery detection. We then present our main findings on improved \nameref{sec:classification}, leading up to the goals listed above. As a consistency check, we also discuss the authentication of synthetic forgeries created by Stable Diffusion and StyleGANs \nameref{sec:synthetic_detection}. A brief summary is provided in the \nameref{sec:conclusion}.

\section*{Methodology}\label{sec:methods}
In this section, we provide an overview of the methods we employed to generate synthetic images for art authentication. We first outline the process of creating synthetic images for the training datasets and provide details about the composition of the dataset. We also elaborate on our classification methodology for art authentication. Finally, we explain how we evaluate the authentication efficiency.

\subsection*{Methods for synthetic image generation}
\label{sec:generators}

We use two fundamentally different GenAI methods to generate synthetic artwork: an image-to-image generative adversarial network (GAN) and a text-to-image diffusion model. The images generated both by the diffusion model and GAN are collectively referred to as synthetic data.

We used the NVlabs implementation \stgan~\cite{stylegan3} which is one of the most recent and successful GANs.
\stgan\ was trained from scratch on 10380 portraits in various genres and by many different authors, including 126 portraits by van Gogh, 280 by Modigliani, and 157 by Raphael. The portraits by the three artists are sourced from Wikiart, and while there is considerable overlap, they do not entirely represent the sets of original artworks detailed in the subsequent datasets. The latter ones are not limited to portraits but, in turn, are filtered to include only artworks appearing in museum collections or Catalogue Raisonnés,  ensuring a high level of certainty regarding their authenticity.

The training took 5M epochs on 4 GPUs. More details on the training procedure and the quality of the resulting images can be found in the supporting information \nameref{S1_Appendix}. With such training, \stgan\ produces images in a mixture of styles by random authors. We used the trained  \stgan\ to produce a ``raw'' dataset of 2000 random portraits. In what follows, images picked at random from this ``raw'' dataset are referred to as the ``raw GANs'' image set. Furthermore, subsets of synthetic portrait images in the style of a specific artist (van Gogh, Modigliani, or Raphael) were created through further training for 50k epochs exclusively on original paintings by the respective artist. This yielded image sets of synthetic images that looked stylistically close to the works of van Gogh, Modigliani, and Raphael. We refer to these datasets as ``tuned GANs'' image sets.
We remark that, due to the limited number of paintings available for each artist, prolonged training on the exclusive data sets often results in a  decline in the quality of the \stgan\ images. Rather than achieving the desired outcome of generating a large variety of images in a given style, long specialized training tends to produce an almost exact reproduction of the training set. Specialized training for some time between 20k and 100k epochs has proven to be a good compromise, striking a balance between a wide variety of images and effective adaption of the desired style.

To create the text-to-image synthetic artworks, we use the Stable Diffusion~\cite{Rombach22} generative model. It relies on CLIP guidance~\cite{CLIP_guidance} to semantically align the latent text representation and the latent image representation and a U-Net architecture~\cite{U-Net2015} as a de-noising diffusion model. The quality of images generated using Stable Diffusion strongly depends on the text prompt. We generate images in the style of each artist using a simple prompt indicating the style, the content, and the artist; for example: `Post-impressionist painting of a young boy, by Vincent van Gogh'. We adopt the Stable Diffusion version 2.1 (\texttt{v2-1\_768-ema-pruned.ckpt}), with 60 inference steps, 8 guidance scale, and $512 \times 512$ pixels resolution. The resulting synthetic dataset is referred to as ``diffusion''. Note that Stable Diffusion has been trained on subsets of the very broad open-source dataset \texttt{LAION-2B(en)} collected \textit{in the wild} and using the large contrastive model OpenCLIP while the GAN was trained on the controlled \texttt{WikiArt} dataset. We used the second version of Stable Diffusion because it is trained on fully open data and models.

\subsection*{Composition of training datasets}
\label{sec:dataset}

AI-based art authentication is a binary classification task where the model learns to differentiate between authentic and non-authentic artworks (including known forgeries). This requires training on two sets of artworks for each artist, an authentic and a contrast set. Our experiments are centered around the van Gogh dataset which contains $126$ original artworks from the VGDB-2016 dataset \cite{folego2016vangogh}. The dataset was gathered from Wikimedia Commons, and it contains artworks with a similar chronology or artistic movement to van Gogh and with a density of at least $196.3$ PPI (Pixels Per Image), the dataset also contains two artworks with debated attribution for testing. Here we note that the number of images does not exactly match those mentioned in the original paper \cite{folego2016vangogh}, we provide the number of images that were actually downloaded through the dataset link provided in \cite{folego2016vangogh}.

In addition, in supporting information \nameref{S1_Appendix} we provide two further tests of our approach on the artworks by Modigliani ($100$ original artworks) and Raphael ($206$ original artworks) and imitations/forgeries thereof. The latter datasets were collected from museum collections or sourced from \textit{Catalogue Raisonnées}, which are expert-curated lists documenting all verified authentic artworks by the respective artists.

The contrast set includes artworks that were not made by the artist, but that resemble it closely and are helpful in detecting forgeries of the artist's work. Normally, this includes artworks of similar artists, referred to as `proxies', and forgeries or explicit imitations of the artist, referred to as `imitations`. Proxies are paintings by different human authors who painted in a similar style to the artist (i.e.\ artists pertaining to the same artistic movement) and/or were collaborators, pupils, and teachers. The word imitation is used here as an umbrella term to encompass human-made non-autograph copies of authentic works, artworks explicitly made in the style of the artist, and known forgeries of the artist. To these elements, we add synthetic fakes generated by Stable Diffusion 2.1 and StyleGAN3, and we test whether their addition increases the performance of the models. 

The contrast set of Vincent van Gogh contains $212$ artworks by similar artists, $19$ forgeries ($11$ by Otto Wacker and $8$ by John Myatt), $30$ Stable Diffusion generated images, $30$ GANs fine-tuned on the artist, and $30$ random GANs (the `raw GANs'). The set of `raw GANs' contains the same exact images across all three datasets used in this work. 

All images are pre-processed according to the procedure detailed in \cite{Schaerf2023}. Specifically, we generate sub-images of paintings, i.e., RGB images normalized to a fixed size of $256 \times 256$ pixels, with channel values normalized to the unit interval. These sub-images are created by dividing the entire image into $2^p \times 2^p$ equally sized units, where $p$ is determined by the resolution of the original image. If the smaller side of an image is larger than $1024$ pixels, then $p = 2$; if the smaller side is larger than $512$ pixels and smaller than $1024$, then $p = 1$. For all images, irrespective of resolution, we also include the sub-image obtained by center-cropping a square from the full image.
Therefore, depending on the resolution of the original image, the images are patched in $21$, $5$, or $1$ adjacent non-overlapping patches. Using bi-cubic resampling, we reshape all the images to either $224 \times 224$ pixels or $256 \times 256$ pixels depending on the input supported by the model. 

The patches are split randomly into training (72\%), validation (11\%), and test (17\%) sets, ensuring that patches belonging to the same original image feature in the same set. We randomly sample the split 10 times, obtaining 10 bootstrapped splits for cross-validation.

For the sake of clarity, we will present the results for the van Gogh dataset \cite{VanGoghDataset} in the remainder of this paper. The outcomes for Modigliani and Raphael are available in the supporting information \nameref{S1_Appendix}. 

\begin{figure}[!ht]
\centering
\caption{{\bf Illustration of real (top row) and synthetic (bottom row) van Gogh images.}\\
``Self Portrait with a Straw Hat'', Vincent van Gogh (1887) ~\cite{Bailey_2021} (square-cropped, top left), ``Self-portrait with a Bandaged Ear and Pipe'', sold by Otto Wacker, previously attributed to van Gogh ~\cite{Harvard_2019} (top right), fine-tuned GAN generated image in the style of van Gogh (bottom left), and Stable Diffusion generated image in style of van Gogh (bottom right).
}
    \includegraphics[width=.45\columnwidth,height=.45\columnwidth]{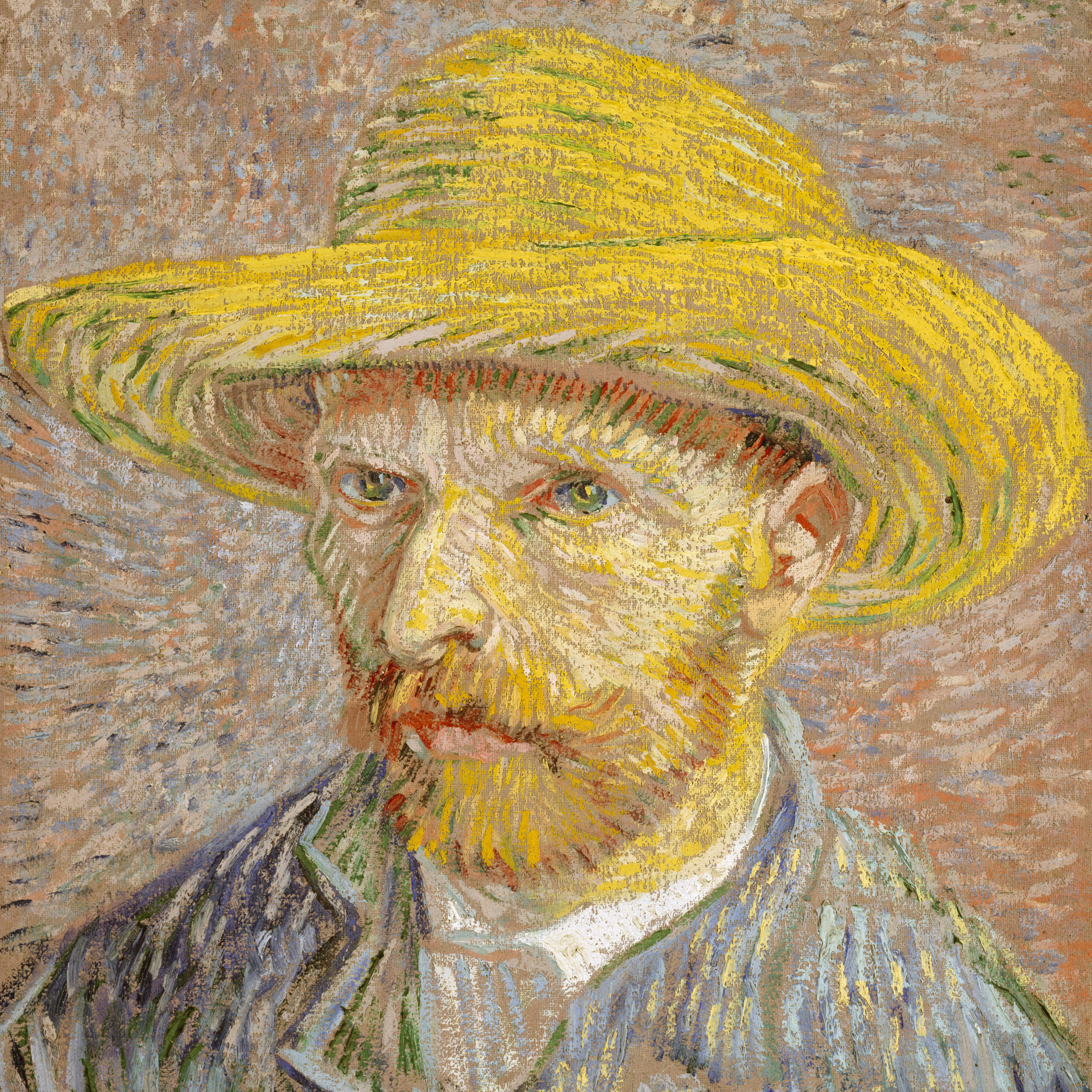}\,%
	\includegraphics[width=.45\columnwidth]{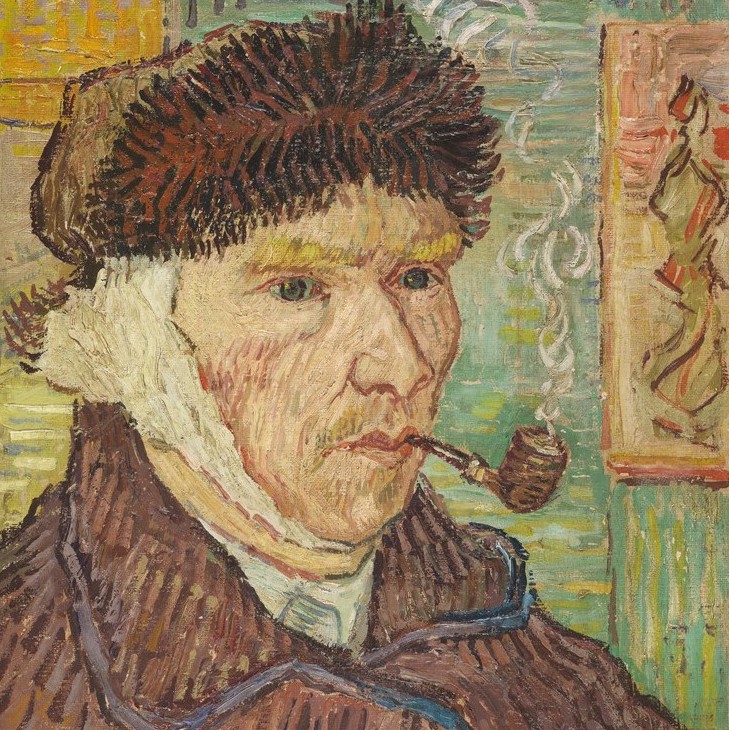}\\
	\includegraphics[width=.45\columnwidth]{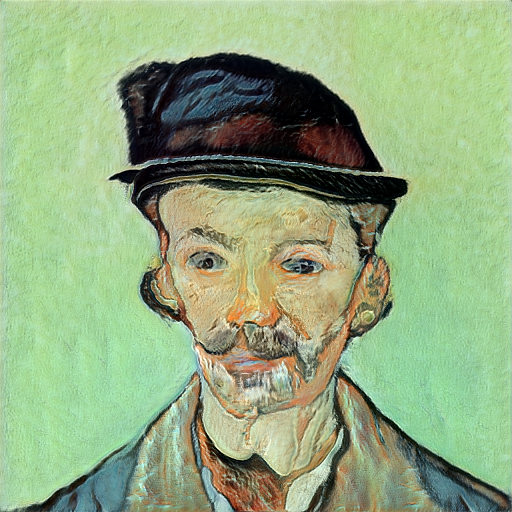}\,%
	\includegraphics[width=.45\columnwidth]{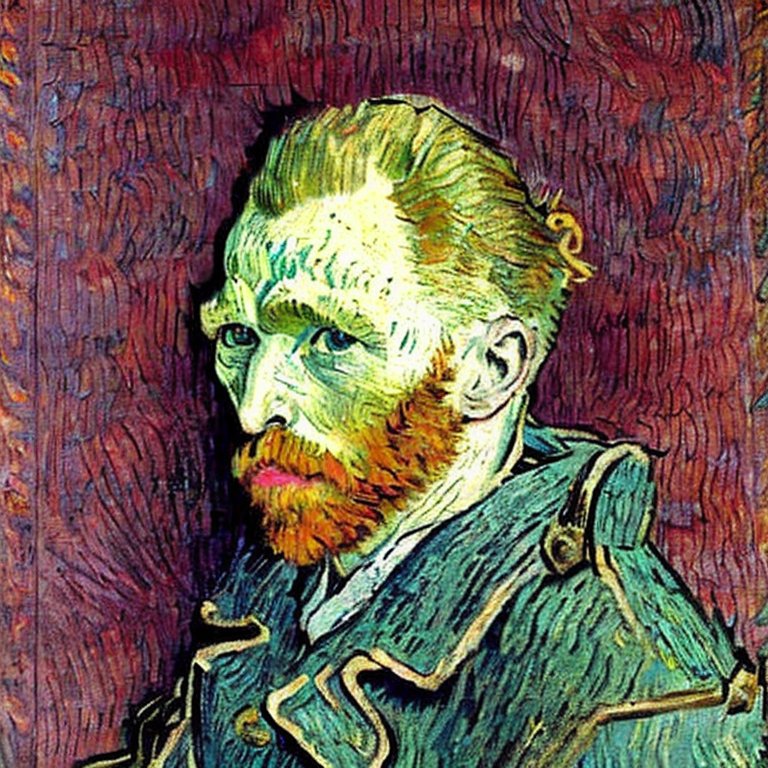}%
	\label{fig1}
\end{figure}

Table~\ref{table1} provides a detailed overview of the van Gogh dataset. The rows represent the six image sets, while the columns show the number of images and the corresponding number of patches. Representative images of each class (authentic, imitation, GAN, and diffusion) are shown in Fig.~\ref{fig1}.
%The names of the contrast sets reflect their contents in terms of constituent image sets. All contrast sets have the same composition of imitations and proxies, but differ in the compositions of synthetic image sets. In both cases, our largest datasets are referred to as ``diffusion + GANs'' and contain post-selected GAN-generated images and images created by Stable Diffusion, along with human-made imitations and proxy paintings.

\begin{table}[!ht]
\centering
\caption{{\bf Composition of the van Gogh dataset.}}
    \begin{tabular}{|l|c|c|} \hline
     \textbf{Image set} & \textbf{Number of images}  & \textbf{Patches} \\ 
		 \hline 
		authentic  & 126 &  2582  \\ \hline
		imitations  & 19 &  271 \\ \hline
		proxies   & 212 & 4208  \\ \hline
		tuned GANs  & 30 & 150  \\ \hline
		raw GANs  & 30 & 150 \\ \hline
		diffusion  & 30 & 150 \\ \hline
	\end{tabular}
\label{table1}
\end{table}

\subsection*{Classification methodology}
\label{sec:classification}

After preparing the training and testing datasets of human-made and synthetic artwork forgeries as described in the previous Section, we proceed to explain the classification methodology on the gathered dataset. In line with the approach outlined in~\cite{Schaerf2023}, we employ transformer-based classification methods (specifically, Swin Base~\cite{liu2021Swin}) and state-of-the-art Convolutional Neural Networks (EfficientNet B0~\cite{pmlr-v97-tan19a}) to distinguish between authentic artworks and forgeries. These models have been adopted in~\cite{Schaerf2023} and proved to outperform the canonical ResNet101 model, the typical baseline model for art authentication. The Swin Base is an image transformer model that uses a hierarchical structure with shifting windows to reduce the computational complexity of transformer models, it accepts inputs of size $224 \times 224$ and has $88M$ parameters. On the other side, EfficientNet B0 is a CNN-based model belonging to the class of EfficientNets which adopts an optimal width, depth, and resolution scaling for the architecture. It accepts slightly larger inputs of shape $256 \times 256$ and has only $5.3M$ parameters. 
We note that Swin Base is a larger model version compared to EfficientNet B0.
We will present the classification outcomes for both the Swin Base and EfficientNet models, however, we do not directly compare the performance of these models. Rather, the purpose is to demonstrate that incorporating synthetic data into training datasets enhances classification reliability, regardless of the classifier architecture.

We use the Swin Base and EfficientNet models pre-trained on ImageNet data~\cite{deng2009imagenet} and fine-tune them for the art authentication task. To do so, we substitute the final activation layer with one dense layer converging in a single node with \textit{sigmoid }activation and train using the \textit{binary cross-entropy} loss without freezing the weights. For both models, we use a learning rate of \num{e-5}, a batch size of $32$. We train the models on binary classification, where class $1$ contains the authentic artworks by the artist (authentic set) and class $0$ refers to the non-authentic artworks (proxies, imitations, and synthetic images). 

To investigate how the addition of synthetic images in the training set improves classification accuracy we run the following experiments. First, we test whether the addition of each of the synthetic sets separately, as well as the combination of Stable Diffusion and fine-tuned GANs, improves the classification accuracy of the human-made forgeries against a baseline trained using `proxies' and `imitations'. This baseline also agrees with the previous work \cite{Schaerf2023}.

\begin{figure}[!ht]
\caption{{\bf Composition of the training and testing sets for the different experiments.}\\Each box in the training represents a training configuration. The configuration names on the bottom row are used throughout the following sections. Green sub-boxes indicate the original set, red indicates the contrast set. }
\includegraphics[width=\columnwidth]{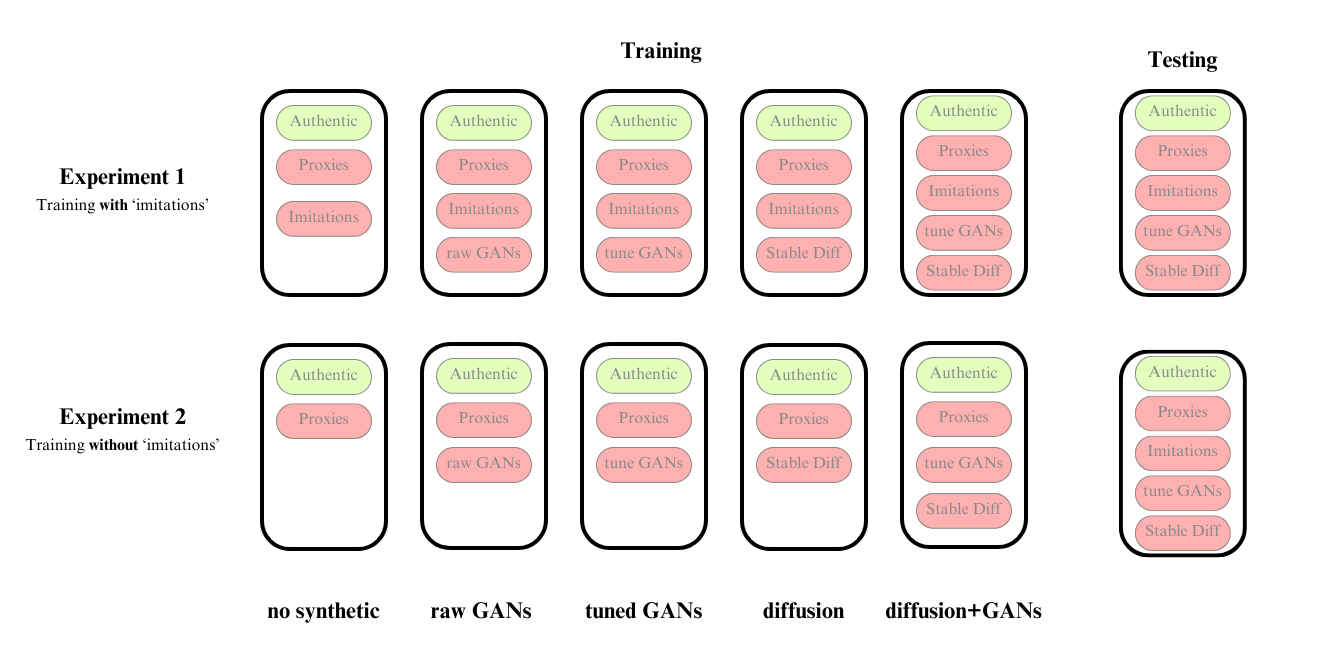}
\label{fig2}
\end{figure}

Secondly, we investigate the extent to which synthetic images can increase the detection of human-made forgeries while never training the models on any human forgeries, thus relying solely on `proxies' and excluding `imitations'. The setup of the experiments is schematized in Fig.~\ref{fig2}. We note that the second task is inherently much harder than the first one, as it tests whether synthetic images \textbf{alone} can substitute the need to train on human-made forgeries to detect such forgeries. This case scenario addresses the situations in which there are no known forgeries of an artist but it is still desirable to be able to flag possible forgeries. This case is common in art connoisseurship as less well-known artists are rarely forged.
To evaluate the performance of our classifiers, we compute the confusion matrices and classification accuracy on the test datasets aggregated at the image level, separately for authentic artworks, imitations, and synthetic images. 

Statistical significance of the results is guaranteed by the use of cross-validation with 10 different splits and subsequent uncertainty estimation. All quoted results are the median of the joint distributions and the uncertainties are the (symmetrized) 68$\%$-quantiles of the median (equivalent to the $1\sigma$-standard-error for normally distributed data). We use the concise parenthesis notation in Tables~\ref{table2}-\ref{table4}. Thus, an entry like $\num{0.710 \pm 0.046}$ means that the central value of the distribution we obtained during the cross-validation process is $\num{0.710}$ and it is unlikely (at most 32$\%$) that the true value does deviates from this central value by more than $\num{0.046}$. Similarly, in Figs.~\ref{fig3} and~\ref{fig4} this would correspond to a main bar at $\num{0.710}$ and error-bars of lengths $\num{0.046}$ both up and down.

\section*{Results}
\label{sec:results}

The outcomes of our classification experiments for van Gogh are shown in Tabs.~\ref{table2} and~\ref{table3}. The results are also visually depicted in Fig.~\ref{fig3}. The evaluation of the classification performance is based on two main criteria: the accuracy in classifying human forgeries (accuracy `forgeries') and the accuracy in classifying authentic paintings (accuracy `originals'). Note that what we here refer to as `forgeries' is synonymous with the set of `imitations', as those imitations are, in this case, indeed forgeries.

\subsection*{Detection of human forgeries}
\label{sec:human_detection}

\begin{figure}[!ht]
\caption{{\bf Accuracies of different models for originals and forgeries.}\\
Based on the results presented in Tables~\ref{table2} and \ref{table3} with the composition of the underlying van Gogh data set as detailed in Table~\ref{table1} and visualised in Fig.~\ref{fig2}. The horizontal dotted line shows the baseline without synthetic images in the training data. Similar results for the artists Modigliani and Raphael can be found in \nameref{S1_Appendix}.
}
\resizebox{0.98\textwidth}{!}{\large
	\includegraphics{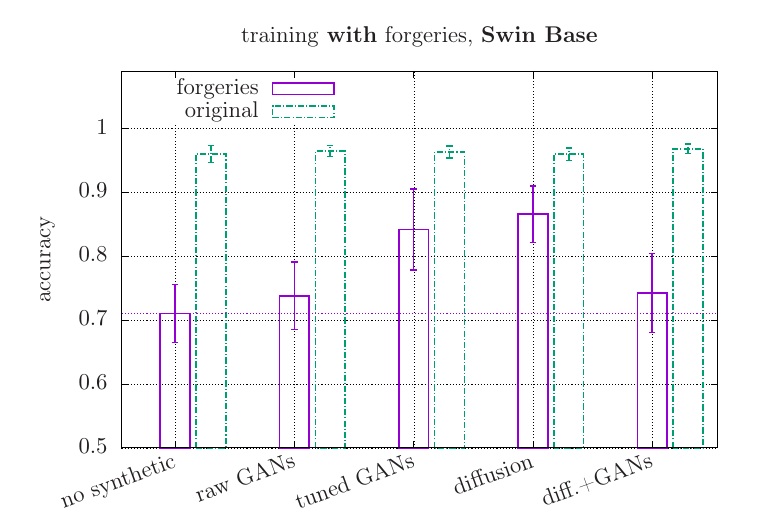}%
	\includegraphics{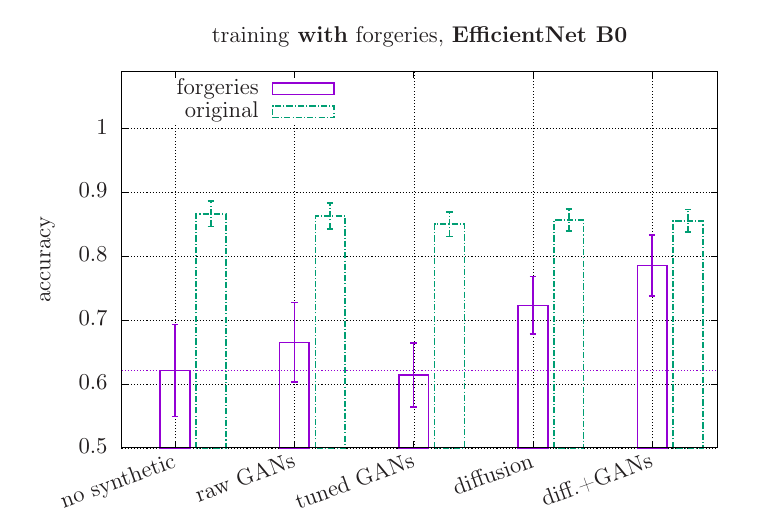}%
}
\resizebox{0.98\textwidth}{!}{\large
	\includegraphics{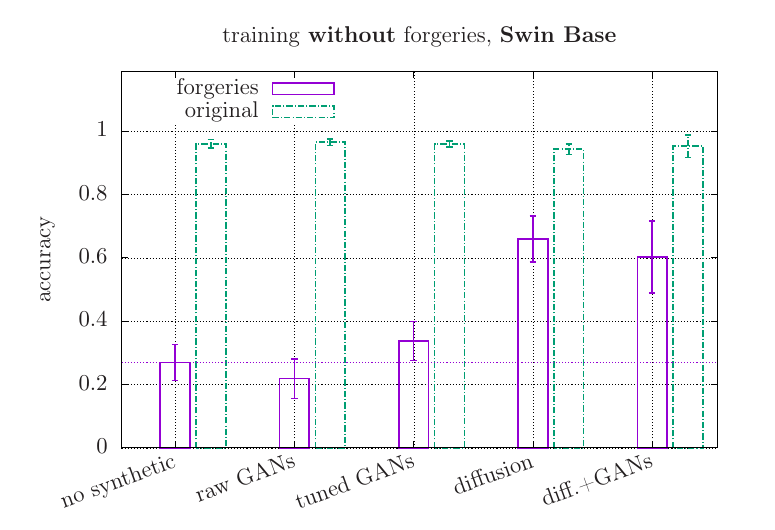}%
	\includegraphics{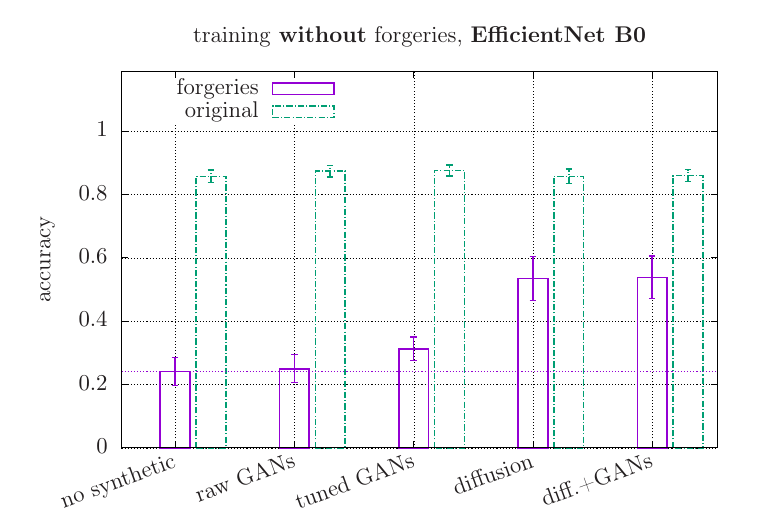}%
}
\label{fig3}
\end{figure}

The classification accuracy for authentic paintings reveals consistently high levels, approximately 90$\%$ or higher with the `Swin Base' model classifier and at least 80$\%$ with `EfficientNet B0', across all training sets, as indicated by the green bars in Fig.~\ref{fig3}.
We observe, moreover, the reproducible improvement in the classification accuracy of human-made forgeries when synthetic forgeries are added to the training datasets. This finding is, to the best of our knowledge, yet unobserved in the literature. On the mixed synthetic training datasets, we were able to achieve accuracies approaching 80$\%$ (see Table~\ref{table2}). Images generated by Stable Diffusion appear to be particularly beneficial, leading to accuracy improvements of 10$\%$ to 20$\%$. These improvements are evident in Fig.~\ref{fig3}, where the purple bars (forgeries) associated with ``no synthetic'' values (this baseline is also extended as a dotted horizontal line) are consistently lower or equal to the values associated with the synthetic (``diffusion'' and ``tuned GANs'') counterparts. This result is particularly impressive considering that the forgeries were often painted by professionals with the goal of avoiding detection.

In addition to augmenting the human-made forgeries with synthetic ones, we also investigated the case where no human-made forgeries were included in the contrast set at all (experiment 2). All classification accuracies are bound to be lower in this case, which is what we observe, and we certainly cannot recommend using this approach in practice if any human-made forgeries are available. However, it allows to resolve the benefit of synthetic forgeries with higher statistical significance. As can be seen in Table~\ref{table3} and the lower two panels of Fig.~\ref{fig3}, the addition of synthetic data allowed to improve the forgery detection accuracy by almost 40$\%$ and 30$\%$ for `Swin Base' and `EfficientNet B0', respectively, both corresponding to a significance of about $4\sigma$.

Finally, the quality of synthetic data plays a crucial role in training success, as expected. As seen in Tables~\ref{table2} and~\ref{table3}, training solely on ``raw GAN'' images resulted in minor or no improvement in authentication accuracy.
Nevertheless, it is interesting to note that in some cases, the addition of ``raw GAN'' datasets without any author-specific features led to slight enhancements in authentication capabilities.
Consistent with previous findings by Schaerf et al.~\cite{Schaerf2023} and the inherent differences in model sizes, the transformer-based Swin Base classifier demonstrated slightly superior overall performance.

All findings are consistent across the two models `Swin Base' and `EfficientNet B0'. This observation is also supported by a similar analysis of Modigliani and Raphael's datasets described in supporting information \nameref{S1_Appendix}. While the numerical values of the classification results may vary between models and artists, the qualitative conclusions remain consistent across all six combined cases.

\begin{table}[!ht]
\centering
\caption{
{\bf Performance on different tests after training with forgeries.}\\
 The composition of the underlying van Gogh data set is detailed in Table~\ref{table1} and visualised in Fig.~\ref{fig2}. The best result for each test is highlighted in \textbf{bold}. Values are medians with respective uncertainties in parentheses.}
    \begin{tabular}{|l|c||c|c|} \hline
    training         & model           & accuracy        & accuracy \\
    contrast set     & architecture    & forgeries      & originals \\ \hline\hline
    no~synthetic	&	Swin~Base	&	$\num{0.710 \pm 0.046}$	&	$\num{0.960 \pm 0.013}$	\\ \hline
    no~synthetic	&	EfficientNet~B0	&	$\num{0.621 \pm 0.072}$	&	$\num{0.867 \pm 0.020}$	\\ \hline
    raw~GANs	&	Swin~Base	&	$\num{0.738 \pm 0.053}$	&	$\num{0.965 \pm 0.009}$	\\ \hline
    raw~GANs	&	EfficientNet~B0	&	$\num{0.665 \pm 0.062}$	&	$\num{0.863 \pm 0.020}$	\\ \hline
    tuned~GANs	&	Swin~Base	&	$\num{0.842 \pm 0.064}$	&	$\num{0.963 \pm 0.010}$	\\ \hline
    tuned~GANs	&	EfficientNet~B0	&	$\num{0.614 \pm 0.050}$	&	$\num{0.850 \pm 0.019}$	\\ \hline
    diffusion	&	Swin~Base	&	\pmb{$\num{0.866 \pm 0.044}$}	&	$\num{0.960 \pm 0.010}$	\\ \hline
    diffusion	&	EfficientNet~B0	&	$\num{0.723 \pm 0.045}$	&	$\num{0.857 \pm 0.017}$	\\ \hline
    diffusion+GANs	&	Swin~Base	&	$\num{0.742 \pm 0.062}$	&	\pmb{$\num{0.968 \pm 0.007}$}	\\ \hline
    diffusion+GANs	&	EfficientNet~B0	&	$\num{0.785 \pm 0.048}$	&	$\num{0.855 \pm 0.018}$	\\ \hline
    \end{tabular}
    \label{table2}
\end{table}

\begin{table}[!ht]
	\centering
	\caption{
		{\bf Performance on different tests after training without forgeries.}\\
		The composition of the underlying van Gogh data set is detailed in Table~\ref{table1} and visualised in Fig.~\ref{fig2}. The best result for each test is highlighted in \textbf{bold}. Values are medians with respective uncertainties in parentheses.
	}
	\begin{tabular}{|l|c||c|c|} \hline
		training         & model           & accuracy        & accuracy \\
		contrast set     & architecture    & forgeries      & originals \\ \hline\hline
		no~synthetic	&	Swin~Base	&	$\num{0.269 \pm 0.057}$	&	$\num{0.960 \pm 0.014}$	\\ \hline
		no~synthetic	&	EfficientNet~B0	&	$\num{0.241 \pm 0.044}$	&	$\num{0.857 \pm 0.020}$	\\ \hline
		raw~GANs	&	Swin~Base	&	$\num{0.219 \pm 0.063}$	&	\pmb{$\num{0.965 \pm 0.010}$}	\\ \hline
		raw~GANs	&	EfficientNet~B0	&	$\num{0.250 \pm 0.044}$	&	$\num{0.873 \pm 0.018}$	\\ \hline
		tuned~GANs	&	Swin~Base	&	$\num{0.337 \pm 0.062}$	&	$\num{0.959 \pm 0.009}$	\\ \hline
		tuned~GANs	&	EfficientNet~B0	&	$\num{0.313 \pm 0.037}$	&	$\num{0.876 \pm 0.017}$	\\ \hline
		diffusion	&	Swin~Base	&	\pmb{$\num{0.660 \pm 0.073}$}	&	$\num{0.943 \pm 0.017}$	\\ \hline
		diffusion	&	EfficientNet~B0	&	$\num{0.535 \pm 0.069}$	&	$\num{0.857 \pm 0.023}$	\\ \hline
		diffusion+GANs	&	Swin~Base	&	$\num{0.603 \pm 0.114}$	&	$\num{0.953 \pm 0.036}$	\\ \hline
		diffusion+GANs	&	EfficientNet~B0	&	$\num{0.538 \pm 0.068}$	&	$\num{0.860 \pm 0.018}$	\\ \hline
	\end{tabular}
	\label{table3}
	%\end{adjustwidth}
\end{table}

\subsection*{Detection of synthetic images}
\label{sec:synthetic_detection}

Machine learning methods for the detection of synthetic artwork and synthetic images is currently a very active research topic (see e.g. \cite{Islam_2020_CVPR,Gragnaniello2021,Ojha2023TowardsUF,Guo_2023_CVPR}). While the primary focus of this paper is the detection of art forgeries created by humans, in this Section we demonstrate that, in agreement with previous studies, our classifier Neural Networks (Swin Base and EfficientNet) are also capable of detecting synthetic artwork forgeries created by GenAI. A novel aspect of our approach is that, unlike in most previous studies, the classifier is trained on both human-made and synthetic forgeries. 

\begin{figure}[!h]
\caption{{\bf Accuracies of different models for synthetic data.}\\
	Based on the results shown in Table~\ref{table4} with the composition of the underlying van Gogh data set as detailed in Table~\ref{table1} and visualised in Fig.~\ref{fig2}. 
}
\resizebox{0.98\textwidth}{!}{\large
	\includegraphics{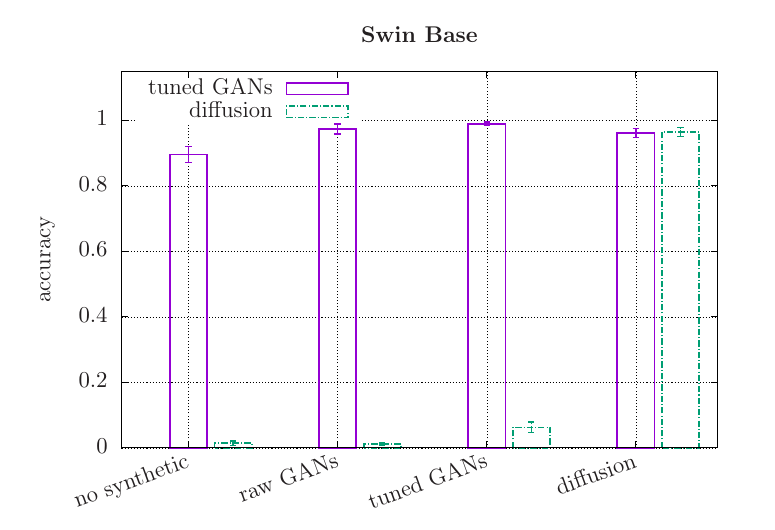}%
	\includegraphics{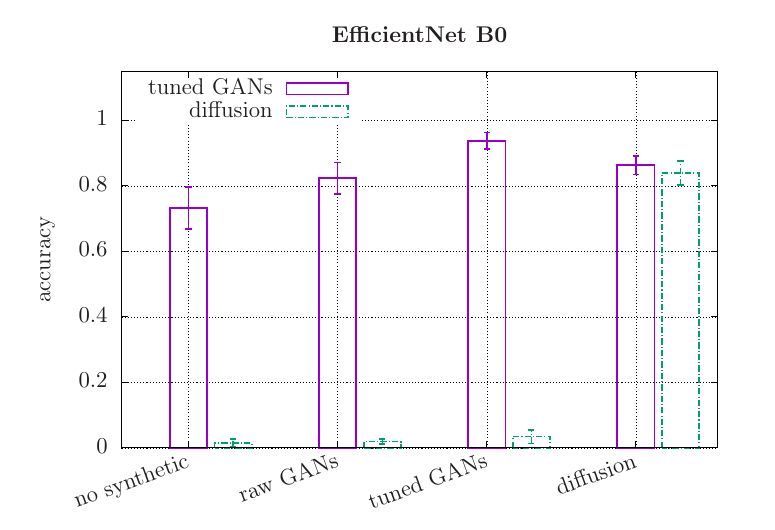}%
}
\label{fig4}
\end{figure}

We assess the efficiency of the detection of synthetic forgeries using the synthetic sets listed in Table~\ref{table1}. Tuned GANs and Stable Diffusion images are tested independently with central values (medians) and uncertainties of the cross-validation results are computed in the same way as in teh previous Section \nameref{sec:human_detection}. The results are summarized in Table~\ref{table4} and in Fig.~\ref{fig4}, which show a considerable improvement of synthetic images detection when integrating synthetic images in the training set. 

Our findings are consistent with the typical conclusion in literature (e.g.~\cite{synthetic-detection,Wang2020,Gragnaniello2021,Ojha2023TowardsUF,Guo_2023_CVPR}), in that the training on synthetic artwork (tuned GANS, diffusion and raw GANs) is crucial for the classifier to detect forgeries created by GenAI.

We have to distinguish the two cases here of training the classifier with a similar GenAI as has to be detected versus training with a different one.
In agreement with the literature, the highest authentication accuracy is obtained if the classifier had already seen synthetic images by the same generator architecture during the training~\cite{Ojha2023TowardsUF, Guo_2023_CVPR}. As one can infer from Table~\ref{table4}, the best results (all above $80\%$) for tuned GANs detection are achieved when tuned GANs are also included in the training and equivalently training on Stable Diffusion images allows the highest accuracy for diffusion detection. We also observe that including tuned GANs in the training helps to some extent with the detection of images generated by Stable Diffusion, and vice versa. This is an interesting observation given that most previous studies on synthetic forgery detection concentrated on generator-specific image features and visual inconsistencies~\cite{Wang2020,lighting_inconsistency,perspective_inconsistency}.

In the van Gogh based studies presented in this main manuscript it turned out that our classifier could detect GAN images relatively well even without training on synthetic data. The trends described above are visible for both, tuned GANs and diffusion, but they are significantly more pronounced for the latter. When performing the same analysis with the artists Modigliani and Raphael (see \nameref{S1_Appendix}), we found that in some cases tuned GANs also eluded detection very effectively (some accuracies below $10\%$) as long as no \stgan\ images had been included in the training. In all of these cases training on the given architecture readily improved the accuracy.

\begin{table}[!ht]
%\begin{adjustwidth}{-2.25in}{0in}
\centering
\caption{
{\bf Accuracy of synthetic forgery detection.}\\
 The composition of the underlying van Gogh data set is detailed in Table~\ref{table1} and visualised in Fig.~\ref{fig2}. The best result for each test is highlighted in \textbf{bold}. Values are medians with respective uncertainties in parentheses.
}
    \begin{tabular}{|l|c||c|c|} \hline
    training        & model           & accuracy        & accuracy        \\
    contrast set    & architecture    & \text{Stable}       & \text{tuned}       \\
                    &                 & \text{Diffusion}      & \text{GANs}         \\
    \hline \hline                     
    no~synthetic	&	Swin~Base	&	$\num{0.014 \pm 0.007}$	&	$\num{0.895 \pm 0.025}$	\\ \hline
    no~synthetic	&	EfficientNet~B0	&	$\num{0.015 \pm 0.013}$	&	$\num{0.732 \pm 0.064}$	\\ \hline
    raw~GANs	&	Swin~Base	&	$\num{0.011 \pm 0.003}$	&	$\num{0.972 \pm 0.015}$	\\ \hline
    raw~GANs	&	EfficientNet~B0	&	$\num{0.019 \pm 0.008}$	&	$\num{0.823 \pm 0.048}$	\\ \hline
    tuned~GANs	&	Swin~Base	&	$\num{0.062 \pm 0.016}$	&	\pmb{$\num{0.989 \pm 0.005}$}	\\ \hline
    tuned~GANs	&	EfficientNet~B0	&	$\num{0.034 \pm 0.020}$	&	$\num{0.937 \pm 0.025}$	\\ \hline
    diffusion	&	Swin~Base	&	\pmb{$\num{0.964 \pm 0.014}$}	&	$\num{0.961 \pm 0.013}$	\\ \hline
    diffusion	&	EfficientNet~B0	&	$\num{0.838 \pm 0.037}$	&	$\num{0.862 \pm 0.028}$	\\ \hline
%    diffusion+GANs	&	Swin~Base	&	$\num{0.942 \pm 0.037}$	&	$\num{0.981 \pm 0.008}$	\\ \hline
%    diffusion+GANs	&	EfficientNet~B0	&	$\num{0.834 \pm 0.038}$	&	$\num{0.952 \pm 0.005}$	\\ \hline                                   
    \end{tabular}                             
    \label{table4}%
%    \end{adjustwidth}
\end{table}

\section*{Discussion and Conclusions}
\label{sec:conclusion}

In this work, we demonstrated that additional training on synthetic forgeries can help classifier neural networks to detect human-made art forgeries.
To this end, we generated synthetic images using \texttt{Stable Diffusion 2.1}~\cite{Rombach22} as well as \stgan~\cite{stylegan3}, and we added these images to our training contrast sets. We discovered that the introduction of synthetic data into the training set significantly improves the accuracy of the detection of human-made forgeries. This holds for all the artists and classifier architectures we tested, both with GAN-generated and Stable Diffusion images, though the latter is clearly preferable. The results are particularly significant when the training set does not include any human-made forgeries at all. This result is particularly novel as it hints at the possibility that synthetic forgeries are similar enough to human forgeries to be able to teach something to the classifier. 

In agreement with previous research, we also confirmed that the training on synthetic images expectably improves the authentication efficiency on synthetic forgeries, especially when the same GenAI architecture was used to produce the training dataset.

Further exploration should be dedicated to quantifying the optimal ratio of human-made forgeries to synthetic data. Moreover, it might be of interest to investigate the influence of image resolutions on the performance of both generators and classifiers. However, significantly larger computing resources than currently available to us would be needed for this type of analysis. With more computational resources, an additional improvement to this work might be the dedicated post-training of generators like Stable Diffusion on authentic artworks of given artists in order to further enhance the quality of synthetic data.

A notable limitation of our study is that the synthetic GAN-based forgeries are limited to portrait paintings due to the poor convergence of other types of images. While the artistic styles of van Gogh, Modigliani and Raphael are without doubt very different, further tests should be carried out to generalize the findings to a variety of genres and even more different artists.

\section*{Acknowledgments}
We thank Romanas Einikis for his valuable support with all the technical details of the computation, and Johanna Induni for the classification of images into genres. Further thanks go to Simon Hands for his help with initiating the project.
This work was funded in part by the STFC Impact Acceleration Award to the University of Liverpool.
Numerical simulations were undertaken on Barkla, part of the High Performance Computing facilities at the University of Liverpool, UK.
%\nolinenumbers

%\bibliographystyle{vancouver}
%\bibliographystyle{plos2015}
\bibliography{bibliography}

\appendix

\section*{Supporting information}
\paragraph*{S1 Appendix.}
\label{S1_Appendix}~\\

{\bf 1. Image generation using StyleGAN.} 
This section presents details regarding the training configurations employed by StyleGAN3. Additionally, we list the numbers of images used for training, along with the corresponding quality of the generated results, sorted by categories. Furthermore, we include visual representations of sample images generated after the training.

{\bf 2. Classification results for Modigliani and Raphael.}
The entire workflow presented for the artist Vincent van Gogh in the main manuscript has also been performed for Amedeo Modigliani and Raphael (Raffaello Sanzio da Urbino). The results can be found in this section.

\section{Evaluation of the quality of synthetic images generated by \texttt{StyleGAN}}

In this section of the supplementary information to the manuscript ``Synthetic images aid the recognition of human-made art forgeries'' we detail the procedure employed for the training of \stgan~\cite{stylegan3}, the generation of images as well as their quality.

The training was performed independently on genre-based subsets of Wikiart ({\url{www.wikiart.org}), with the number of images of each subset and the corresponding genre listed in Tab.~\ref{tab:data_sets}. We trained each genre starting from white noise (i.e.\ no pre-training) with a resolution of $256\times 256$ pixels. The portraits analysed in the main manuscript were trained in an independent additional run using a higher resolution of $512\times 512$.
	The corresponding hyperparameters we used for the training with \stgan\ are listed in Tab.~\ref{tab:hyperparams}.
	
	\begin{table}[ht]
		\caption{{\bf Hyperparameters used for the training with \stgan~\cite{stylegan3}.}\\
			The full command run for training would read \texttt{python train.py --cfg=stylegan2 --kimag=5000 <hyperparameters>}. We also used \texttt{--snap=50 --tick=1 --metrics==kid50k\_full,fid50k\_full} to monitor the progress.}\label{tab:hyperparams}
		\begin{tabular}{|l|l|l|l|l|l|l|}
			\hline
			resolution & \texttt{--batch=} & \texttt{--gamma=} & \texttt{--cbase=} & \texttt{--glr=} & \texttt{--dlr=} & \texttt{--mbstd-group=} \\\hline\hline
			$256\times 256$ & 16 & 1 & 16384 & $\num{0.001}$ & $\num{0.001}$ & 4 \\\hline
			$512\times 512$ & 12 & 5 &  & $\num{0.001}$ & $\num{0.001}$ & 3 \\\hline
		\end{tabular}
	\end{table}
	
	It is interesting to note that the latest alias-free version included in \stgan\ turned out to perform worse on artworks than the older version \texttt{StyleGAN2}~\cite{stylegan2}, in our case realized by the corresponding flag natively provided in \stgan. This is likely a consequence of local hard transitions, often featured by brush strokes, which tend to be smeared out by the translationally invariant \stgan. 
	
	\begin{table}[ht]
		%\begin{adjustwidth}{-2.25in}{0in} 
		\caption{
			{\bf Training data sets.}\\
			Number of images with resolution at least $256\times 256$, and quality of the training results using \stgan. We provide the Fréchet Inception Distance (FID) as a metric for the quality of the generated images (lower is better).
		}\label{tab:data_sets}
		\centering
		\begin{tabular}{|l|S[table-format=6]|S[round-mode = figures,round-precision = 3]|}
			\hline
			\textbf{Type} & \text{\textbf{No.\ images}}	& \text{\textbf{FID}} \\\hline\hline 
			portraits and self-portraits	&	10380	&	13.58	\\\hline 
			abstract	&	2571	&	42.81	\\                 \hline 
			animals	&	863	&	93.72	\\                      \hline 
			cityscapes	&	3211	&	18.53	\\                  \hline 
			figurative and allegorical	&	2694	&	40.38	\\ \hline 
			history and genre paintings	&	10557	&	14.00	\\ \hline 
			illustrations and sketches	&	880	&	78.46	\\ \hline 
			landscapes	&	10795	&	10.20	\\              \hline
			nude paintings	&	1147	&	63.00	\\              \hline 
			religious paintings	&	4663	&	21.04	\\      \hline 
			still and flower paintings	&	2729	&	39.40	\\ 	\hline
		\end{tabular}	
		%\end{adjustwidth}
	\end{table}
	
	Tab.~\ref{tab:data_sets} provides an overview over the number of images used for training and the resulting image qualities achieved by the GAN for different image types with a resolution of $256\times 256$. The Fréchet Inception Distance (FID) is the state of the art estimator for image quality that is closest to human perception as a rule. A high FID (ca.~20 or more) usually signifies bad results, while a low FID (here less than 20) indicates that the images are reasonably realistic. However, this rule has notable exceptions, in our case `history and genre paintings' as well as `still and flower paintings'. Most humans would readily agree that the former category produced unsatisfactory results (see fig.~\ref{fig:some_images}, top left) while the latter succeeded with the flowers at least (see fig.~\ref{fig:some_images}, bottom right), both contrary to the FID predictions.
	
	\begin{figure}[htb]
		\centering
		\caption{
			{\bf Representative $256\times 256$ images.}\\
			Categories used are `history and genre paintings' (top left), `landscapes' (top right), `figurative and allegorical' (bottom left), and `still and flower paintings' (bottom right).
		}
		\includegraphics[width=.35\columnwidth]{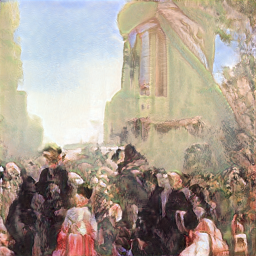}\,%
		\includegraphics[width=.35\columnwidth]{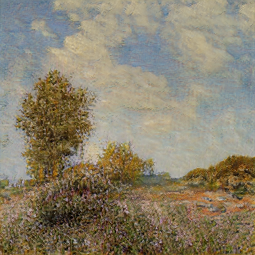}\\
		\includegraphics[width=.35\columnwidth]{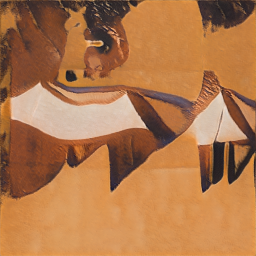}\,%
		\includegraphics[width=.35\columnwidth]{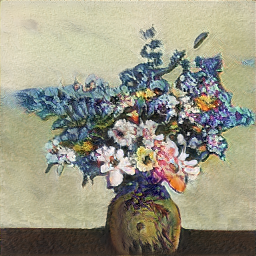}%
		\label{fig:some_images}
	\end{figure}
	
	There is an overall trend that a larger training set results in higher quality images as could have been expected. However, even with similar sample sizes, some categories fare much better than others. Some examples are shown in Fig.~\ref{fig:some_images} where the images in the top row (`history and genre paintings' and `landscapes') have training data sets of similar size. The data sets `figurative and allegorical' and `still and flower paintings' are also similar in size. Their representatives in the bottom row have extremely different quality as well.
	
	We speculate on several causes which can, at least partially, be responsible for such differences in generative quality. One possibility and known issue of GANs is a poor convergence of the optimizer during the training process. A high number of minuscule details is certainly prohibitive when learning on such low resolutions, for instance, many people in the historical paintings, each with facial features. Both, the image resolution and the network capacities are insufficient to resolve this kind of details. In addition, a large diversity of images poses a difficulty because the GAN might not be able to identify reoccurring features and is incapable of generalizations. This is most likely the pivotal problem with the figurative and allegorical paintings.
	
	A separate training run has been performed on images with the higher resolution of $512 \times 512$ for the `portraits and self-portraits` category, which serves as a base for all the analysis in the main Manuscript. These GAN images are used for the benchmarks below.  These are trained on a smaller training set (only $\num{7983}$ of the $\num{10380}$ images had a sufficient resolution) and a higher number of network parameters ($\num{59259432}$ instead of $\num{48768547}$ parameters in total). 
	
	%\FloatBarrier
	\section{Classification results for Modigliani and Raphael}
	
	This section of the supplementary information to the manuscript ``Synthetic images aid the recognition of human-made art forgeries'' complements the results presented in the manuscript for the artist Vincent van Gogh with analogous studies using portraits by Amedeo Modigliani and Raphael (Raffaello Sanzio da Urbino). The entire procedure (from data generation to classification and analysis) is identical with that employed for van Gogh and we refer to the main manuscript for the details.
	
	\subsection*{Data sets}
	
	The compositions of the training datasets are listed in Tabs.~\ref{table:modigliani_compositiondetail} and~\ref{table:raphael_compositiondetail} with representative images of each category displayed in Figs.~\ref{fig:modigliani_images} and~\ref{fig:raphael_images} for Modigliani and Raphael, respectively.
	
	Modigliani’s dataset contains $100$ original artworks, $58$ ’proxies’, $21$ ’imitations’, and the same amount of synthetic fakes as van Gogh, $30$ per type. Raphael, on the other side, contains $206$ original artworks, $96$ ’proxies’, and $99$ ’imitations’, also with the same number of synthetic images.
	
	All displayed images of human-made paintings~\cite{nudeOriginal,SoutineFake,MadonnaSolly,YoungManRed} are in the public domain.
	
	\begin{table}[!ht]
		\centering
		\caption{{\bf Composition of the Modigliani dataset.}}
		\label{table:modigliani_compositiondetail}
		\begin{tabular}{|l|c|c|} \hline
			\textbf{Image set} & \textbf{Number of images}  & \textbf{Patches} \\ 
			\hline 
			authentic  & 100 &  1812  \\ \hline
			imitations  & 21 &  269 \\ \hline
			proxies   & 58 & 1160  \\ \hline
			tuned GANs  & 30 & 150  \\ \hline
			raw GANs  & 30 & 150 \\ \hline
			diffusion  & 30 & 150 \\ \hline
		\end{tabular}
	\end{table}
	
	\begin{table}[!ht]
		\centering
		\caption{{\bf Composition of the Raphael dataset.}}
		\label{table:raphael_compositiondetail}
		\begin{tabular}{|l|c|c|} \hline
			\textbf{Image set} & \textbf{Number of images}  & \textbf{Patches} \\ 
			\hline 
			authentic  & 206 &  2756  \\ \hline
			imitations  & 99 &  1293 \\ \hline
			proxies   & 96 & 1296  \\ \hline
			tuned GANs   & 30 & 150  \\ \hline
			raw GANs  & 30 & 150 \\ \hline
			diffusion  & 30 & 150 \\ \hline
		\end{tabular}
	\end{table}
	
	\begin{figure}[!ht]
		\centering
		\caption{{\bf Illustration of human-made (top row) and synthetic (bottom row) Modigliani images.}\\
			``Nu couché'' (Reclining nude) by Amedeo Modigliani~\cite{nudeOriginal} (square cropped, left), imitation of Modigliani's ``Portrait de Chaïm Soutine''~\cite{SoutineFake,nudeFake} (square cropped, top), fine-tuned GAN generated image in style of Modigliani (bottom left), and Stable Diffusion generated image in style of Modigliani (bottom right).
		}
		\includegraphics[width=.45\columnwidth]{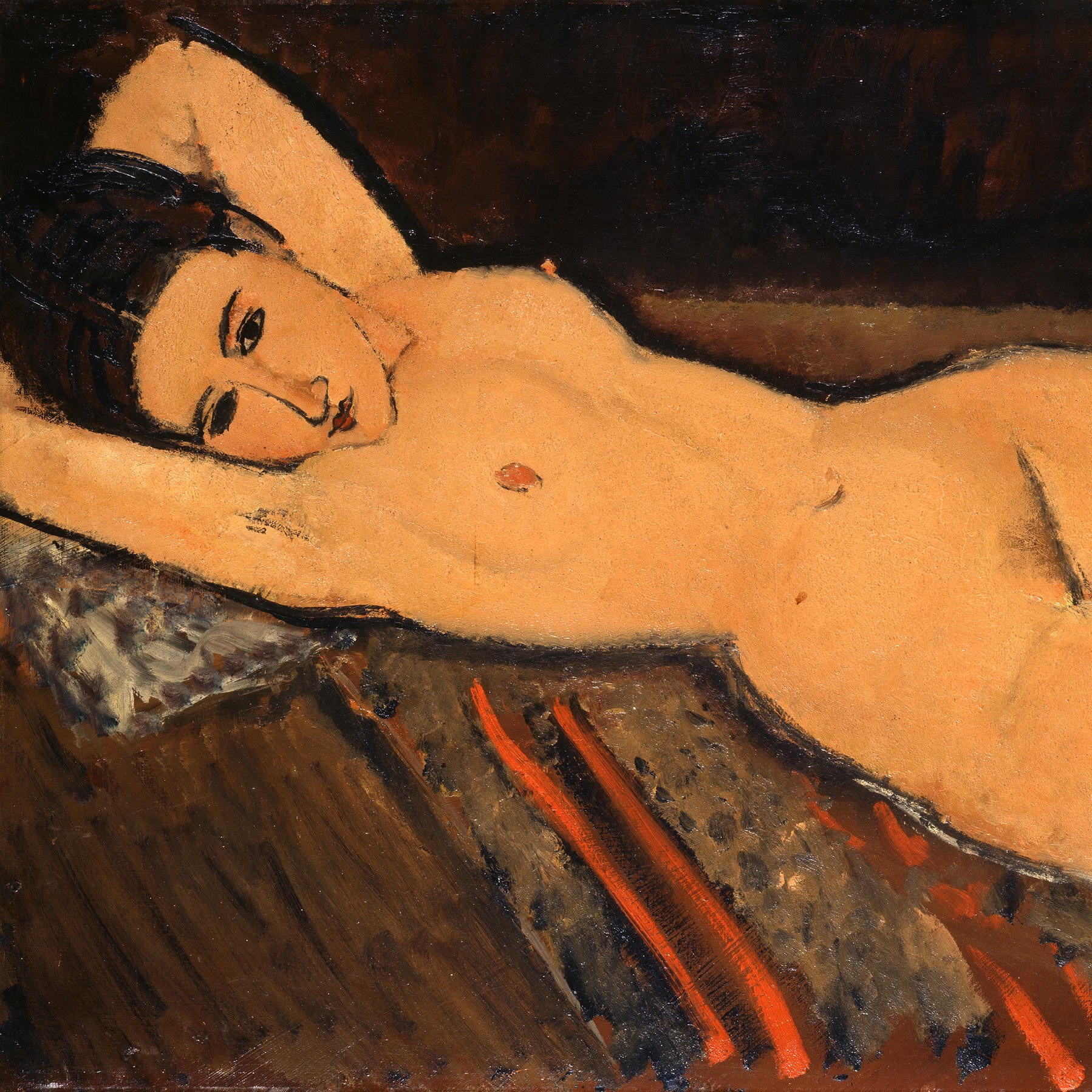}\,%
		\includegraphics[width=.45\columnwidth]{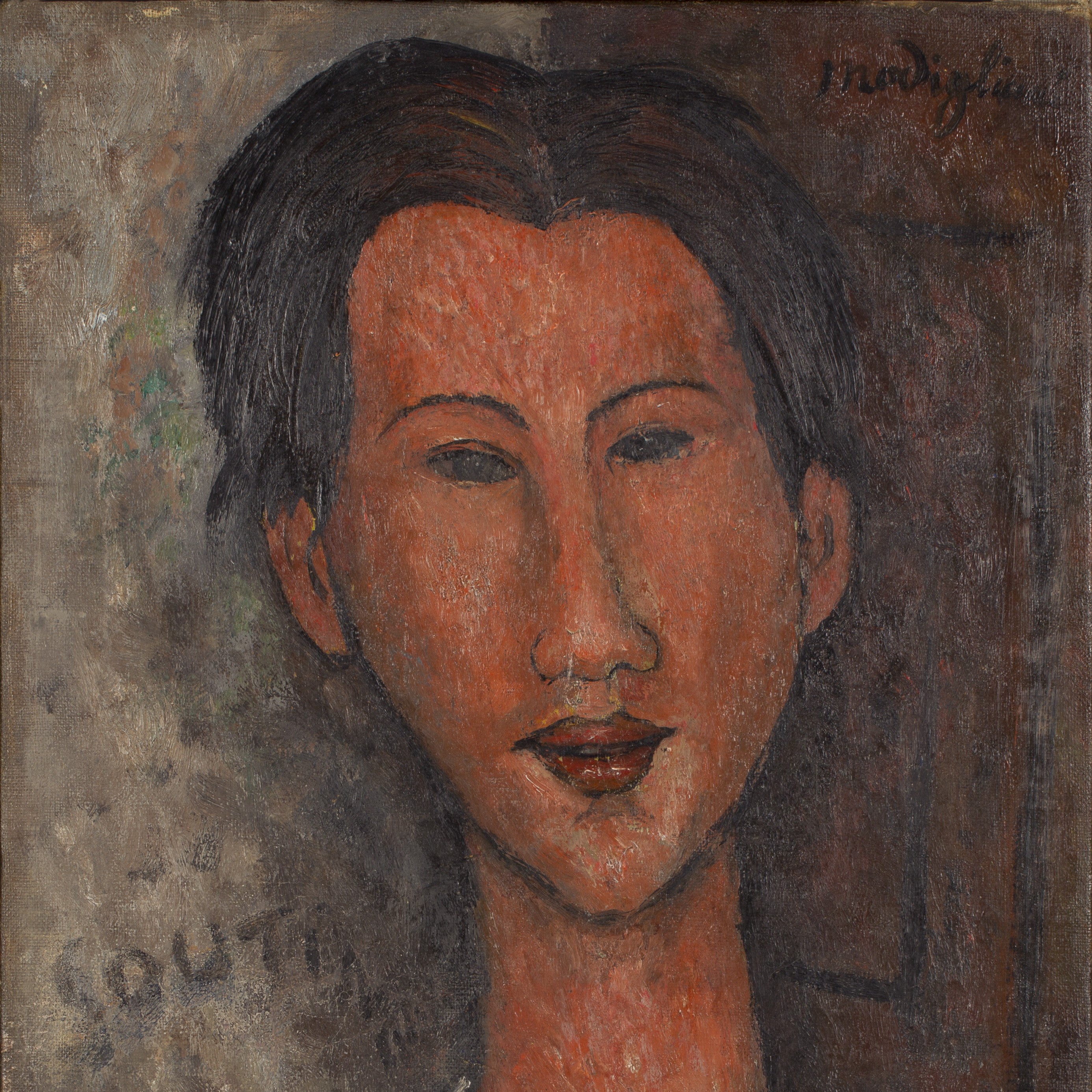}\\
		\includegraphics[width=.45\columnwidth]{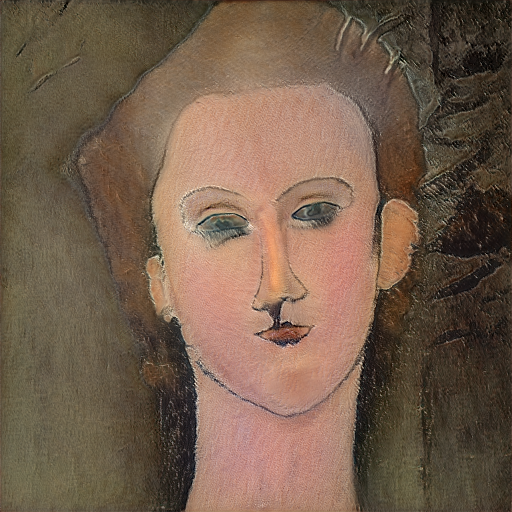}\,%
		\includegraphics[width=.45\columnwidth]{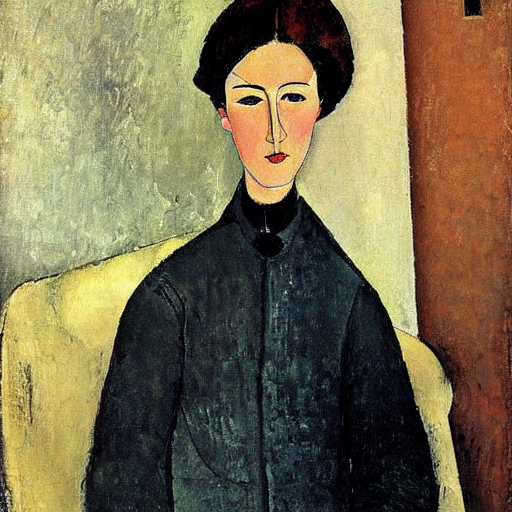}%
		\label{fig:modigliani_images}
	\end{figure}
	
	\begin{figure}[!ht]
		\centering
		\caption{{\bf Illustration of real (top row) and synthetic (bottom row) Raphael images.}\\
			``Madonna with Child'' by Raffaello Sanzio~\cite{MadonnaSolly} (square-cropped, top left), ``Portrait of a Young Man in Red'' by the Circle of Raphael~\cite{YoungManRed} (top right), fine-tuned GAN generated image in the style of Raffaello (bottom left), and Stable Diffusion generated image in the style of Raffaello (bottom right).
		}
		\includegraphics[width=.45\columnwidth]{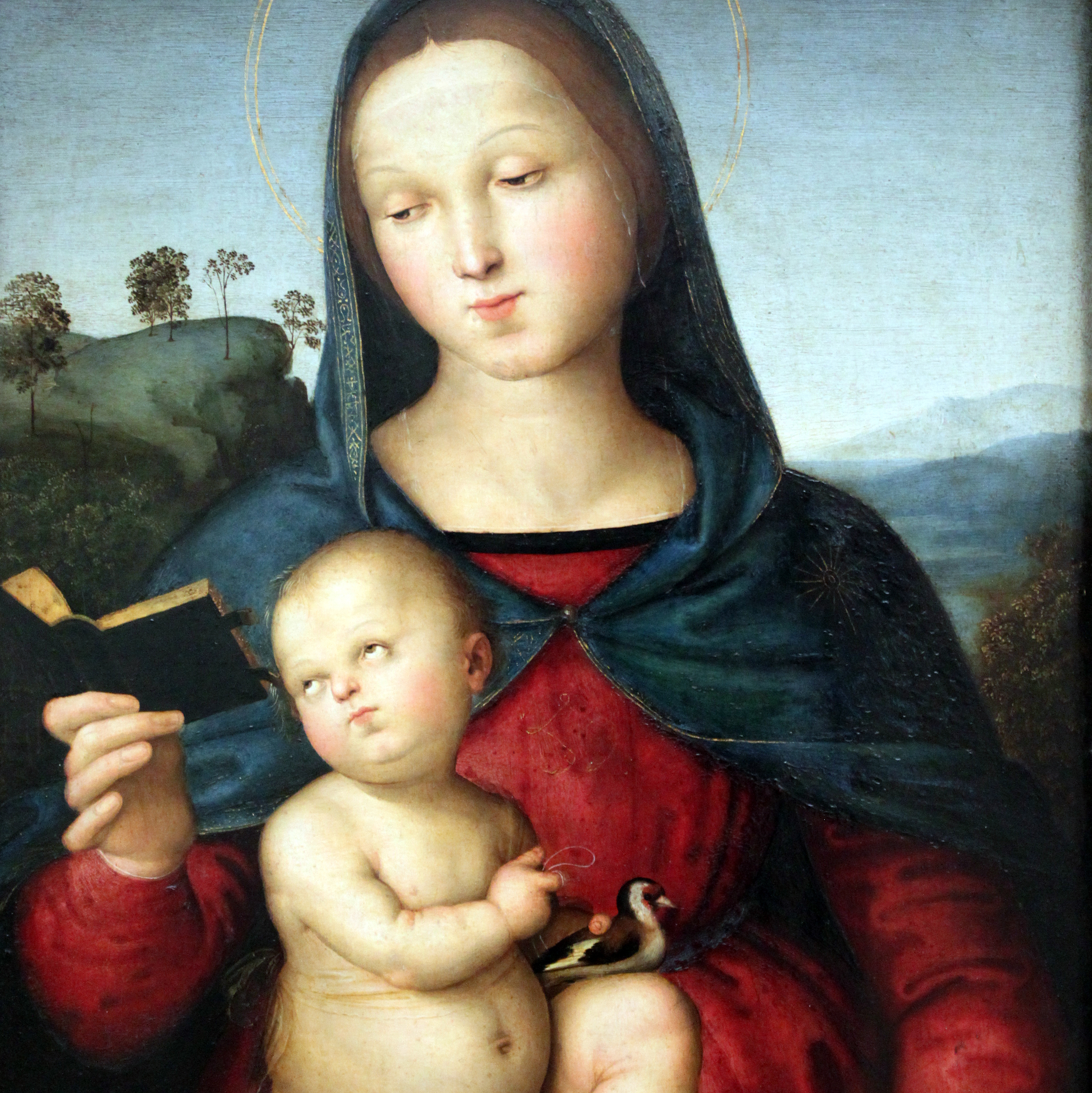}\,%
		\includegraphics[width=.45\columnwidth]{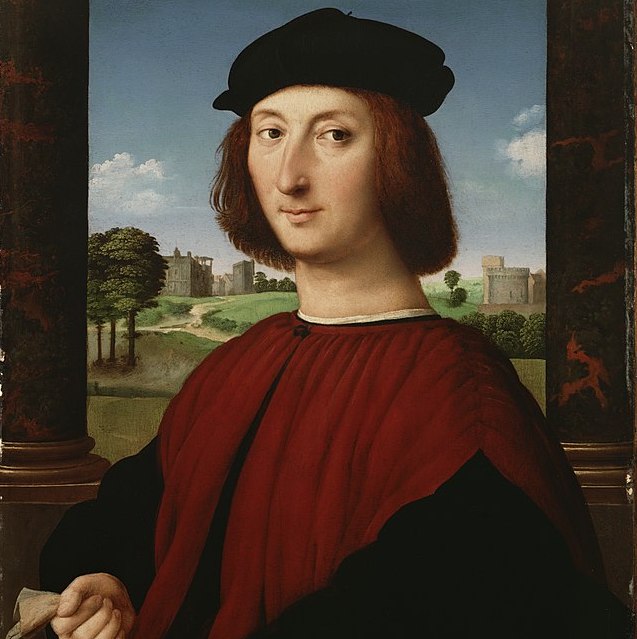}\\
		\includegraphics[width=.45\columnwidth]{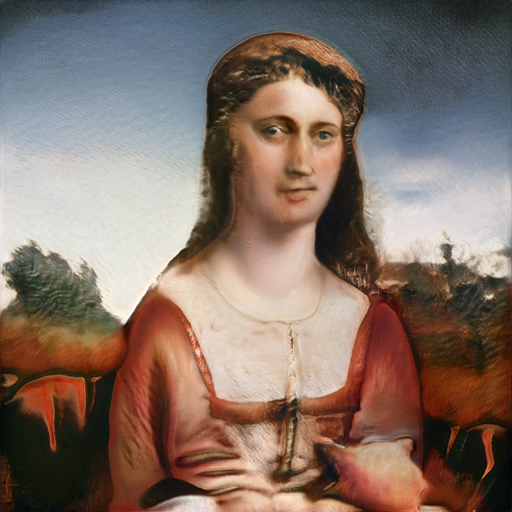}\,%
		\includegraphics[width=.45\columnwidth]{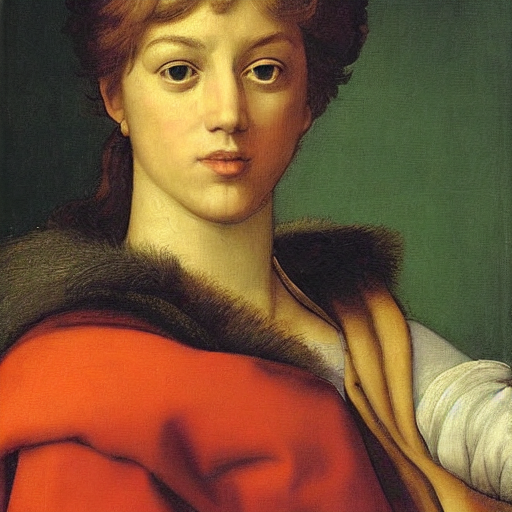}%
		\label{fig:raphael_images}
	\end{figure}
	
	% We no longer split in training and test sets in the cross-validation procedure.
	%\begin{table}[!ht]
	%	\centering
	%	\caption{{\bf Compositions of the Raphael and Modigliani test datasets.}}
	%	\begin{tabular}{|l|c|c|} 
		%		\hline
		%		\textbf{Artist} & \textbf{imitations} & \textbf{authentic} \\ 
		%		\hline
		%		Modigliani  & 13 &  9  \\ \hline
		%		Raphael     & 35 &  20 \\ \hline
		%	\end{tabular}
	%	\label{table:test_compositiondetail}
	%\end{table}
	
	\FloatBarrier
	\subsection*{Detection of human forgeries}
	
	In the following we report the results of the classification experiments: Modigliani with human-made forgeries in the training set (Tab.~\ref{tab:performance_modigliani_forgeries}, top panels of Fig.~\ref{fig:accuracies-modigliani-raphael}), without forgeries (Tab.~\ref{tab:performance_modigliani_noforgeries}, third row of Fig.~\ref{fig:accuracies-modigliani-raphael}), Raphael with forgeries (Tab.~\ref{tab:performance_raphael_forgeries}, second row of Fig.~\ref{fig:accuracies-modigliani-raphael}), without forgeries (Tab.~\ref{tab:performance_raphael_noforgeries}, bottom panels of Fig.~\ref{fig:accuracies-modigliani-raphael}).
	
	The accuracy of the classification of original paintings is consistently high and fully compatible with the results obtained for van Gogh paintings in the main manuscript. The accuracy of forgery classification on the other hand is lower than that observed for van Gogh in most cases, especially when no human-made forgeries are included in the training set. It is important to note that this is in no way contradicting the conclusions drawn in the main manuscript since, regardless of absolute numbers, these accuracies improve significantly in all considered cases when synthetic forgeries are added to the training data.
	
	\begin{table}[!ht]
		%\begin{adjustwidth}{-2.25in}{0in} 
		\centering
		\caption{
			{\bf Performance for Modigliani on different tests after training with forgeries.}
		}
		\begin{tabular}{|l|c||c|c|} \hline
			training         & model           & accuracy        & accuracy \\
			contrast set     & architecture    & forgeries      & originals \\ \hline\hline
			no~synthetic	&	Swin~Base	&	$\num{0.580 \pm 0.101}$	&	$\num{0.932 \pm 0.019}$	\\ \hline
			no~synthetic	&	EfficientNet~B0	&	$\num{0.461 \pm 0.097}$	&	$\num{0.881 \pm 0.009}$	\\ \hline
			raw~GANs	&	Swin~Base	&	$\num{0.561 \pm 0.137}$	&	$\num{0.920 \pm 0.015}$	\\ \hline
			raw~GANs	&	EfficientNet~B0	&	$\num{0.531 \pm 0.118}$	&	$\num{0.869 \pm 0.013}$	\\ \hline
			tuned~GANs	&	Swin~Base	&	$\num{0.491 \pm 0.138}$	&	$\num{0.926 \pm 0.014}$	\\ \hline
			tuned~GANs	&	EfficientNet~B0	&	$\num{0.516 \pm 0.121}$	&	$\num{0.835 \pm 0.015}$	\\ \hline
			diffusion	&	Swin~Base	&	$\num{0.553 \pm 0.117}$	&	$\num{0.932 \pm 0.015}$	\\ \hline
			diffusion	&	EfficientNet~B0	&	$\num{0.484 \pm 0.083}$	&	$\num{0.865 \pm 0.011}$	\\ \hline
			diffusion+GANs	&	Swin~Base	&	$\num{0.605 \pm 0.100}$	&	$\num{0.917 \pm 0.017}$	\\ \hline
			diffusion+GANs	&	EfficientNet~B0	&	$\num{0.543 \pm 0.146}$	&	$\num{0.843 \pm 0.020}$	\\ \hline
		\end{tabular}
		\label{tab:performance_modigliani_forgeries}
		%\end{adjustwidth}
	\end{table}
	
	\begin{table}[!ht]
		%\begin{adjustwidth}{-2.25in}{0in} 
		\centering
		\caption{
			{\bf Performance for Modigliani on different tests after training without forgeries.}
		}
		\begin{tabular}{|l|c||c|c|} \hline
			training         & model           & accuracy        & accuracy \\
			contrast set     & architecture    & forgeries      & originals \\ \hline\hline
			no~synthetic	&	Swin~Base	&	$\num{0.115 \pm 0.051}$	&	$\num{0.998 \pm 0.001}$	\\ \hline
			no~synthetic	&	EfficientNet~B0	&	$\num{0.263 \pm 0.058}$	&	$\num{0.971 \pm 0.004}$	\\ \hline
			raw~GANs	&	Swin~Base	&	$\num{0.153 \pm 0.053}$	&	$\num{0.997 \pm 0.001}$	\\ \hline
			raw~GANs	&	EfficientNet~B0	&	$\num{0.293 \pm 0.060}$	&	$\num{0.961 \pm 0.006}$	\\ \hline
			tuned~GANs	&	Swin~Base	&	$\num{0.193 \pm 0.062}$	&	$\num{0.991 \pm 0.002}$	\\ \hline
			tuned~GANs	&	EfficientNet~B0	&	$\num{0.337 \pm 0.074}$	&	$\num{0.955 \pm 0.007}$	\\ \hline
			diffusion	&	Swin~Base	&	$\num{0.221 \pm 0.049}$	&	$\num{0.992 \pm 0.003}$	\\ \hline
			diffusion	&	EfficientNet~B0	&	$\num{0.326 \pm 0.065}$	&	$\num{0.975 \pm 0.005}$	\\ \hline
			diffusion+GANs	&	Swin~Base	&	$\num{0.262 \pm 0.057}$	&	$\num{0.983 \pm 0.005}$	\\ \hline
			diffusion+GANs	&	EfficientNet~B0	&	$\num{0.422 \pm 0.087}$	&	$\num{0.944 \pm 0.007}$	\\ \hline
		\end{tabular}
		\label{tab:performance_modigliani_noforgeries}
		%\end{adjustwidth}
	\end{table}
	
	\begin{table}[!ht]
		%\begin{adjustwidth}{-2.25in}{0in} 
		\centering
		\caption{
			{\bf Performance for Raphael on different tests after training with forgeries.}
		}
		\begin{tabular}{|l|c||c|c|} \hline
			training         & model           & accuracy        & accuracy \\
			contrast set     & architecture    & forgeries      & originals \\ \hline\hline
			no~synthetic	&	Swin~Base	&	$\num{0.926 \pm 0.026}$	&	$\num{0.804 \pm 0.022}$	\\ \hline
			no~synthetic	&	EfficientNet~B0	&	$\num{0.681 \pm 0.035}$	&	$\num{0.713 \pm 0.016}$	\\ \hline
			raw~GANs	&	Swin~Base	&	$\num{0.902 \pm 0.022}$	&	$\num{0.835 \pm 0.021}$	\\ \hline
			raw~GANs	&	EfficientNet~B0	&	$\num{0.736 \pm 0.033}$	&	$\num{0.718 \pm 0.015}$	\\ \hline
			tuned~GANs	&	Swin~Base	&	$\num{0.906 \pm 0.027}$	&	$\num{0.820 \pm 0.020}$	\\ \hline
			tuned~GANs	&	EfficientNet~B0	&	$\num{0.746 \pm 0.030}$	&	$\num{0.704 \pm 0.018}$	\\ \hline
			diffusion	&	Swin~Base	&	$\num{0.899 \pm 0.022}$	&	$\num{0.838 \pm 0.017}$	\\ \hline
			diffusion	&	EfficientNet~B0	&	$\num{0.701 \pm 0.037}$	&	$\num{0.707 \pm 0.020}$	\\ \hline
			diffusion+GANs	&	Swin~Base	&	$\num{0.924 \pm 0.031}$	&	$\num{0.797 \pm 0.020}$	\\ \hline
			diffusion+GANs	&	EfficientNet~B0	&	$\num{0.719 \pm 0.027}$	&	$\num{0.693 \pm 0.014}$	\\ \hline
		\end{tabular}
		\label{tab:performance_raphael_forgeries}
		%\end{adjustwidth}
	\end{table}
	
	\begin{table}[!ht]
		%\begin{adjustwidth}{-2.25in}{0in} 
		\centering
		\caption{
			{\bf Performance for Raphael on different tests after training without forgeries.}
		}
		\begin{tabular}{|l|c||c|c|} \hline
			training         & model           & accuracy        & accuracy \\
			contrast set     & architecture    & forgeries      & originals \\ \hline\hline
			no~synthetic	&	Swin~Base	&	$\num{0.009 \pm 0.002}$	&	$\num{0.962 \pm 0.009}$	\\ \hline
			no~synthetic	&	EfficientNet~B0	&	$\num{0.083 \pm 0.010}$	&	$\num{0.887 \pm 0.011}$	\\ \hline
			raw~GANs	&	Swin~Base	&	$\num{0.071 \pm 0.011}$	&	$\num{0.940 \pm 0.009}$	\\ \hline
			raw~GANs	&	EfficientNet~B0	&	$\num{0.127 \pm 0.011}$	&	$\num{0.851 \pm 0.015}$	\\ \hline
			tuned~GANs	&	Swin~Base	&	$\num{0.039 \pm 0.008}$	&	$\num{0.950 \pm 0.008}$	\\ \hline
			tuned~GANs	&	EfficientNet~B0	&	$\num{0.111 \pm 0.009}$	&	$\num{0.877 \pm 0.012}$	\\ \hline
			diffusion	&	Swin~Base	&	$\num{0.066 \pm 0.012}$	&	$\num{0.953 \pm 0.008}$	\\ \hline
			diffusion	&	EfficientNet~B0	&	$\num{0.127 \pm 0.012}$	&	$\num{0.848 \pm 0.012}$	\\ \hline
			diffusion+GANs	&	Swin~Base	&	$\num{0.087 \pm 0.013}$	&	$\num{0.950 \pm 0.010}$	\\ \hline
			diffusion+GANs	&	EfficientNet~B0	&	$\num{0.155 \pm 0.016}$	&	$\num{0.829 \pm 0.014}$	\\ \hline
		\end{tabular}
		\label{tab:performance_raphael_noforgeries}
		%\end{adjustwidth}
	\end{table}

	\begin{figure}[!ht]
		\caption{{\bf Accuracies of different models.}\\
			Classification results for Modigliani and Raphael based on \cref{tab:performance_modigliani_forgeries,tab:performance_raphael_forgeries,tab:performance_modigliani_noforgeries,tab:performance_raphael_noforgeries}. The horizontal dotted line shows the baseline without synthetic images in the training data.
		}
		\resizebox{0.98\textwidth}{!}{\large
			\includegraphics{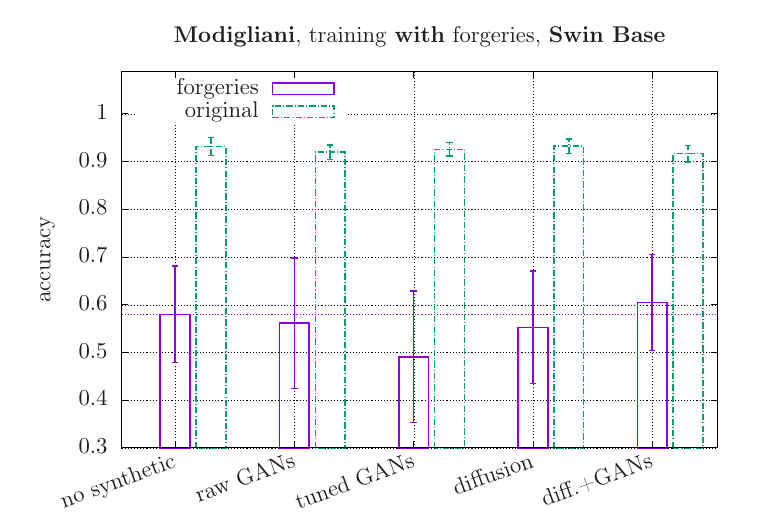}%
			\includegraphics{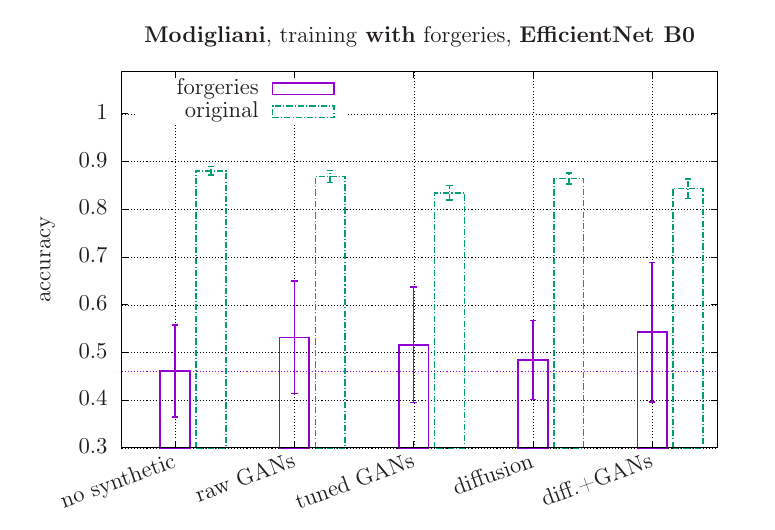}%
		}
		\resizebox{0.98\textwidth}{!}{\large
			\includegraphics{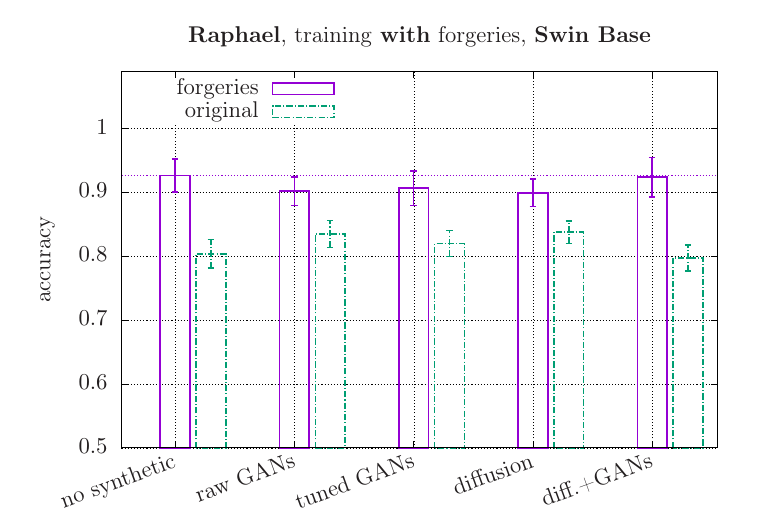}%
			\includegraphics{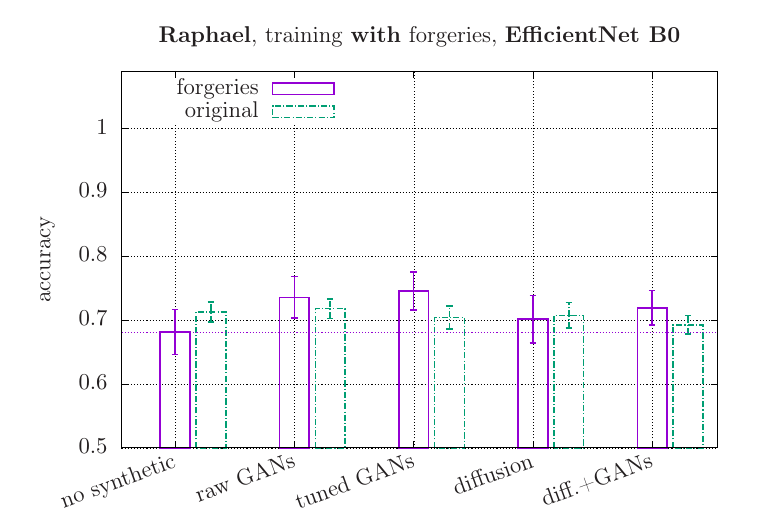}%
		}
		\resizebox{0.98\textwidth}{!}{\large
			\includegraphics{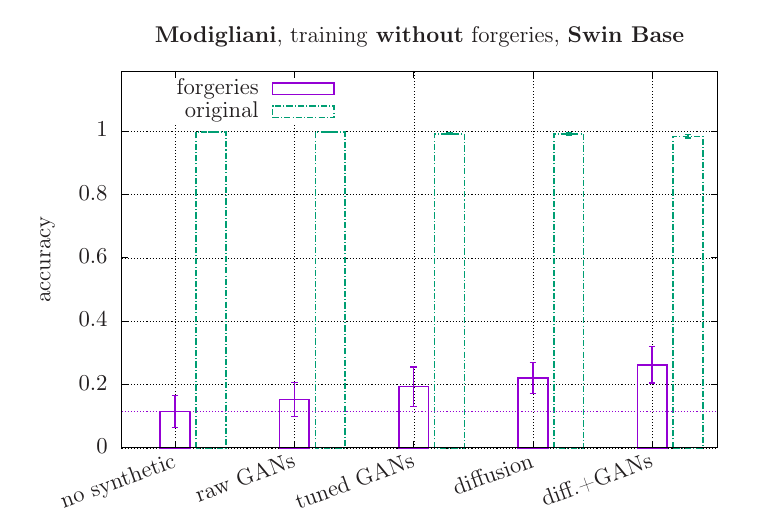}%
			\includegraphics{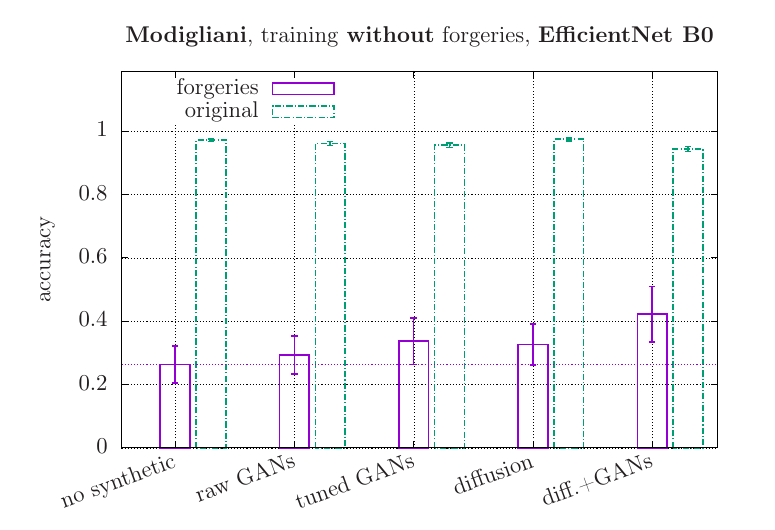}%
		}
		\resizebox{0.98\textwidth}{!}{\large
			\includegraphics{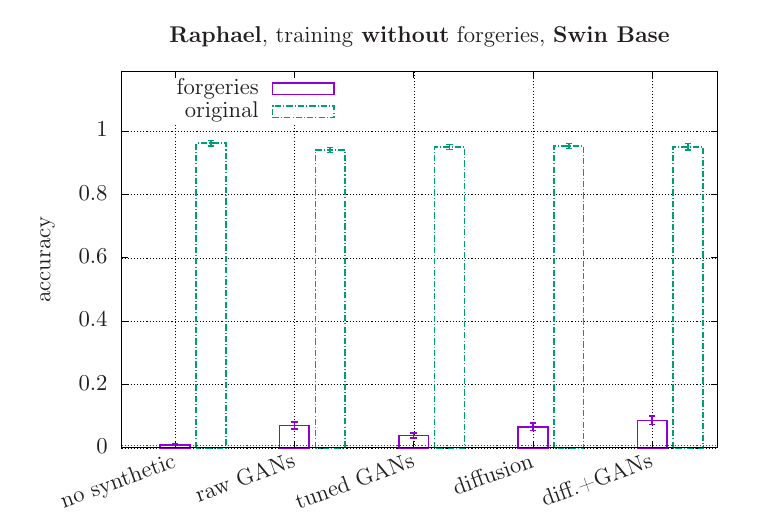}%
			\includegraphics{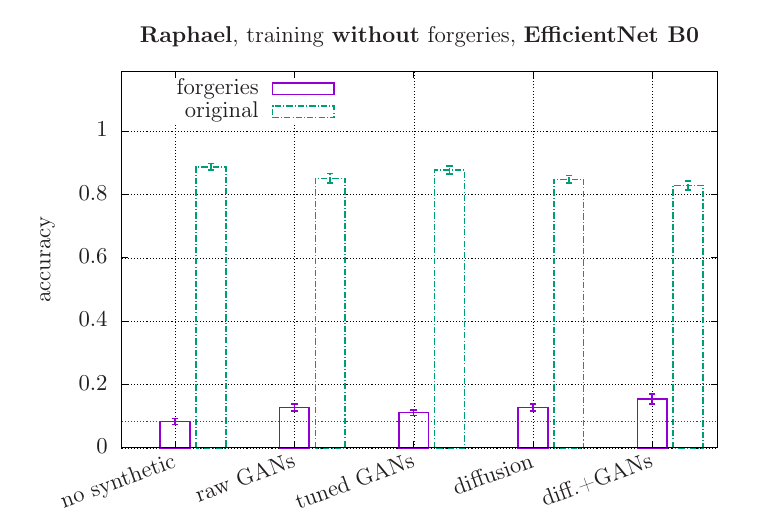}%
		}
		\label{fig:accuracies-modigliani-raphael}
	\end{figure}
	
	\FloatBarrier
	\subsection*{Detection of synthetic images}
	
	As in the main manuscript, we perform a sanity check benchmarking our classifiers on synthetic forgery detections. The results can be found in Tabs.~\ref{tab:performance_synthetic_modigliani} and~\ref{tab:performance_synthetic_raphael} as well as Fig.~\ref{fig:accuracies_synthetic-err}. They consistently support the conclusions that high detection accuracies for a given generator architecture can be obtained when data from this generator is included in the training set and that the detection of adversarial attacks is not reliable when this particular generator is not used for training.
	
	\begin{table}[!ht]
		%\begin{adjustwidth}{-2.25in}{0in}
		\centering
		\caption{
			{\bf Accuracy of synthetic forgery detection for Modigliani.}\\
			%Details of the different contrast sets are presented in Tab.~\ref{table:test_compositiondetail_synthetic}.
		}
		\begin{tabular}{|l|c||c|c|} \hline
			training        & model           & accuracy        & accuracy        \\
			contrast set    & architecture    & \text{Stable}       & \text{tuned}       \\
			&                 & \text{Diffusion}      & \text{GANs}         \\
			\hline \hline                     
			no~synthetic	&	Swin~Base	&	$\num{0.656 \pm 0.093}$	&	$\num{0.027 \pm 0.008}$	\\ \hline
			no~synthetic	&	EfficientNet~B0	&	$\num{0.444 \pm 0.046}$	&	$\num{0.071 \pm 0.013}$	\\ \hline
			raw~GANs	&	Swin~Base	&	$\num{0.790 \pm 0.058}$	&	$\num{0.327 \pm 0.047}$	\\ \hline
			raw~GANs	&	EfficientNet~B0	&	$\num{0.495 \pm 0.068}$	&	$\num{0.177 \pm 0.032}$	\\ \hline
			tuned~GANs	&	Swin~Base	&	$\num{0.792 \pm 0.051}$	&	$\num{0.809 \pm 0.046}$	\\ \hline
			tuned~GANs	&	EfficientNet~B0	&	$\num{0.590 \pm 0.059}$	&	$\num{0.633 \pm 0.045}$	\\ \hline
			diffusion	&	Swin~Base	&	$\num{0.943 \pm 0.020}$	&	$\num{0.045 \pm 0.010}$	\\ \hline
			diffusion	&	EfficientNet~B0	&	$\num{0.832 \pm 0.037}$	&	$\num{0.084 \pm 0.019}$	\\ \hline
			diffusion+GANs	&	Swin~Base	&	$\num{0.980 \pm 0.009}$	&	$\num{0.934 \pm 0.021}$	\\ \hline
			diffusion+GANs	&	EfficientNet~B0	&	$\num{0.892 \pm 0.031}$	&	$\num{0.666 \pm 0.034}$	\\ \hline                                     
		\end{tabular}                             
		\label{tab:performance_synthetic_modigliani}%
		%    \end{adjustwidth}
\end{table}

\begin{table}[!ht]
	%\begin{adjustwidth}{-2.25in}{0in}
	\centering
	\caption{
		{\bf Accuracy of synthetic forgery detection for Raphael.}\\
		%Details of the different contrast sets are presented in Tab.~\ref{table:test_compositiondetail_synthetic}.
	}
	\begin{tabular}{|l|c||c|c|} \hline
		training        & model           & accuracy        & accuracy        \\
		contrast set    & architecture    & \text{Stable}       & \text{tuned}       \\
		&                 & \text{Diffusion}      & \text{GANs}         \\
		\hline \hline                     
		no~synthetic	&	Swin~Base	&	$\num{0.778 \pm 0.059}$	&	$\num{0.322 \pm 0.029}$	\\ \hline
		no~synthetic	&	EfficientNet~B0	&	$\num{0.215 \pm 0.036}$	&	$\num{0.164 \pm 0.019}$	\\ \hline
		raw~GANs	&	Swin~Base	&	$\num{0.781 \pm 0.084}$	&	$\num{0.643 \pm 0.032}$	\\ \hline
		raw~GANs	&	EfficientNet~B0	&	$\num{0.216 \pm 0.034}$	&	$\num{0.279 \pm 0.016}$	\\ \hline
		tuned~GANs	&	Swin~Base	&	$\num{0.841 \pm 0.046}$	&	$\num{0.932 \pm 0.010}$	\\ \hline
		tuned~GANs	&	EfficientNet~B0	&	$\num{0.276 \pm 0.032}$	&	$\num{0.599 \pm 0.011}$	\\ \hline
		diffusion	&	Swin~Base	&	$\num{0.955 \pm 0.024}$	&	$\num{0.387 \pm 0.045}$	\\ \hline
		diffusion	&	EfficientNet~B0	&	$\num{0.539 \pm 0.033}$	&	$\num{0.225 \pm 0.015}$	\\ \hline
		diffusion+GANs	&	Swin~Base	&	$\num{0.971 \pm 0.009}$	&	$\num{0.957 \pm 0.030}$	\\ \hline
		diffusion+GANs	&	EfficientNet~B0	&	$\num{0.544 \pm 0.033}$	&	$\num{0.605 \pm 0.023}$	\\ \hline                                     
	\end{tabular}                             
	\label{tab:performance_synthetic_raphael}%
	%    \end{adjustwidth}
	\end{table}

	\begin{figure}[!h]
\caption{{\bf Accuracies of different models for synthetic data.}\\
	Based on the results shown in~\cref{tab:performance_synthetic_modigliani,tab:performance_synthetic_raphael}. 
}
\resizebox{0.98\textwidth}{!}{\large
	\includegraphics{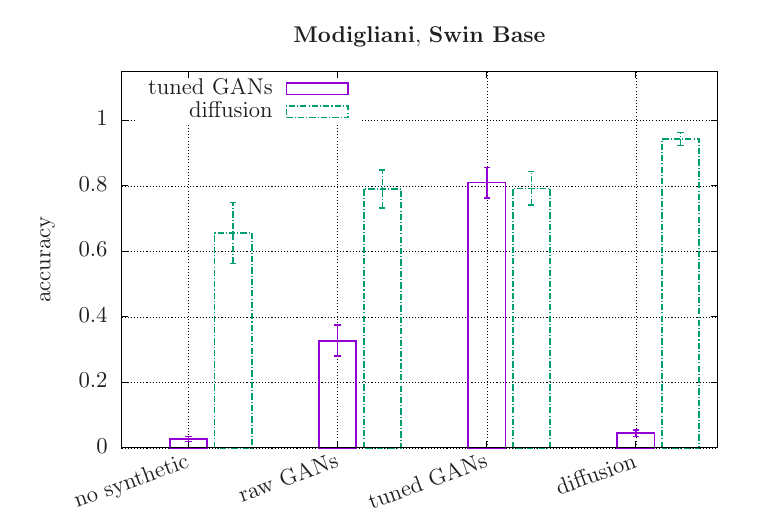}%
	\includegraphics{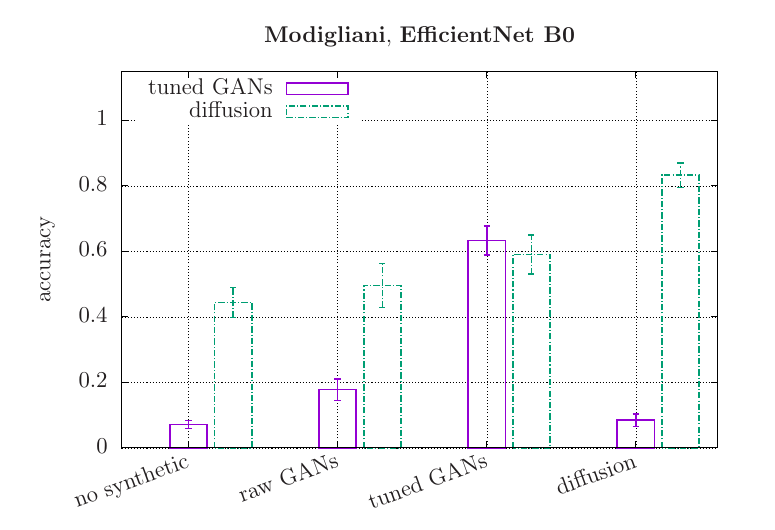}%
}
\resizebox{0.98\textwidth}{!}{\large
	\includegraphics{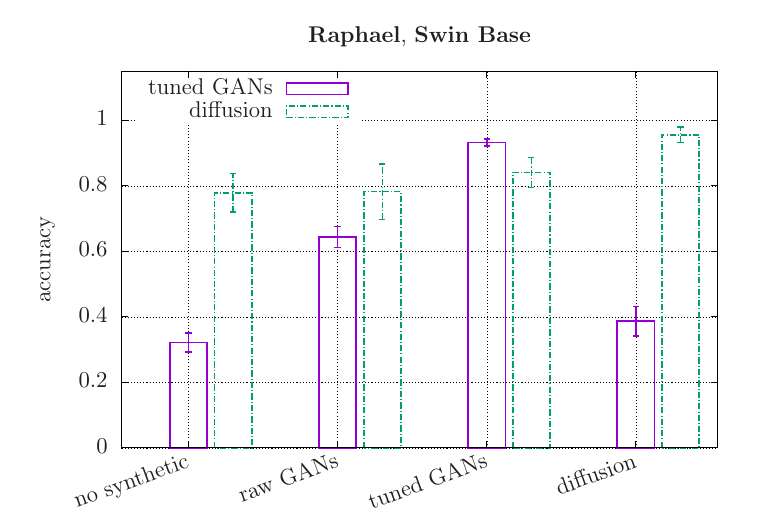}%
	\includegraphics{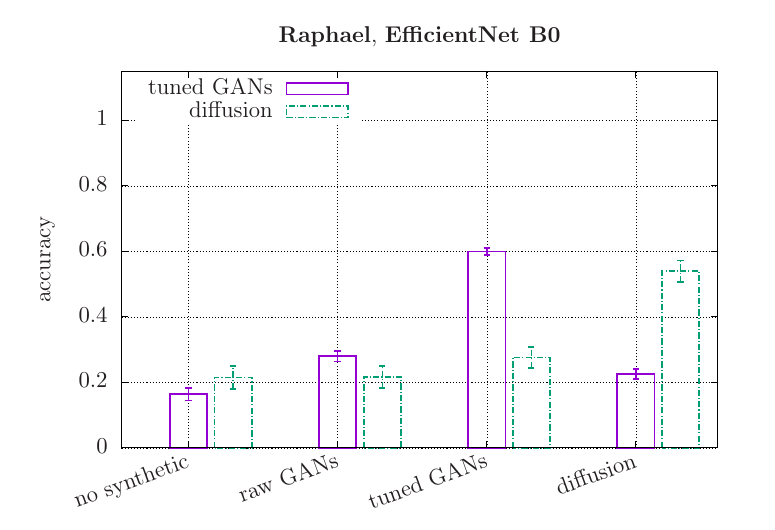}%
}
\label{fig:accuracies_synthetic-err}
\end{figure}

%\FloatBarrier
%\bibliography{bibliography}

\end{document}